\documentclass[10pt,letterpaper,twocolumn]{article}

\usepackage[margin=0.75in]{geometry}
\usepackage[T1]{fontenc}
\usepackage{amsmath}
\usepackage{amssymb}
\usepackage{newtxtext,newtxmath}
\usepackage{microtype}
\usepackage[hyphens]{url}
\usepackage{graphicx}
\usepackage{booktabs}
\usepackage{caption}
\usepackage[numbers,sort&compress]{natbib}
\usepackage{xcolor}
\usepackage[colorlinks=true,
            linkcolor=blue!55!black,
            citecolor=green!35!black,
            urlcolor=blue!65!black]{hyperref}
\hypersetup{
  pdftitle={Teleopit: A Full-Embodiment Humanoid Teleoperation System},
  pdfauthor={Bingqian Wu, Zicheng Xu, Xianghui Fan, Dayu Li, Xiangru Huang}
}

\graphicspath{{figures/}}

\newcommand{\method}{Teleopit}

\title{Teleopit: A Full-Embodiment Humanoid Teleoperation System}
\author{%
  Bingqian Wu\textsuperscript{1,2},
  Zicheng Xu\textsuperscript{1},
  Xianghui Fan\textsuperscript{1},
  Dayu Li\textsuperscript{1},
  Xiangru Huang\textsuperscript{1}\thanks{Corresponding author.}\\[0.4em]
  \normalsize \textsuperscript{1}Westlake University\\
  \normalsize \textsuperscript{2}Shanghai Innovation Institute
}
\date{}

\makeatletter
\let\teleopit@article@maketitle\@maketitle
\renewcommand{\@maketitle}{%
  \teleopit@article@maketitle
  \vspace{-1.5em}
  \begin{center}
    \captionsetup{hypcap=false}
    \includegraphics[width=\textwidth]{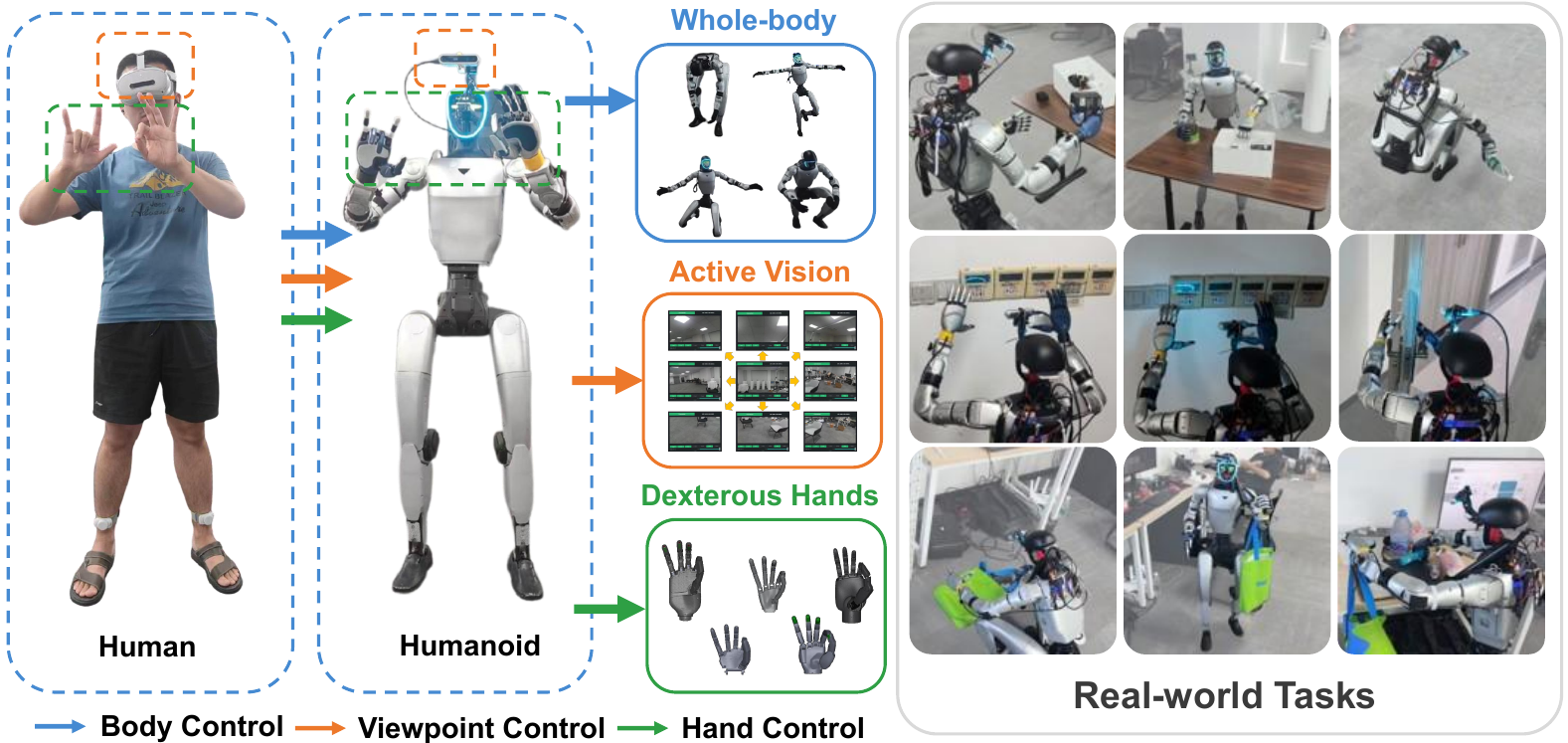}
    \captionof{figure}{\textbf{Teleopit uses VR body, hand, and head tracking
      for full-embodiment humanoid teleoperation.} The interface coordinates
      dynamically feasible whole-body control, configurable dexterous hands,
      and active vision during real-world loco-manipulation.}
    \label{fig:system-teaser}
  \end{center}
  \begingroup
    \@twocolumnfalse
    \begin{abstract}
      Humanoid teleoperation for demonstration collection requires coordinated
      whole-body motion, continuous dexterous hand control, and viewpoint
      control. Existing systems either simplify hand commands or depend on
      dedicated wearable sensors for fine-grained hand motion. We introduce
      \method{}, a full-embodiment teleoperation system that maps body, hand,
      and head signals from VR to a humanoid body, configurable dexterous
      hands, and a 2-DoF active vision module. A history encoder and
      failure-aware rewind sampling improve the motion tracker on both
      motion-capture and live VR references. An optimization-based hand
      retargeter combines normalized finger directions, fingertip closure, and
      thumb-frame alignment to map human hand motion to different dexterous
      hands without tuning hand-specific objective or solver hyperparameters.
      Component experiments evaluate tracking success rate and retargeting
      behavior, while real-robot teleoperation demonstrates coordinated
      locomotion, manipulation, and viewpoint control. ACT and GR00T N1.7
      policies trained on 96 successful demonstrations collected with
      \method{} achieve task success rates of 90.0\% and 95.0\%,
      respectively, when deployed on the humanoid.
      The project page is available at
      \url{https://botrunner64.github.io/teleopit-page}.
    \end{abstract}
  \endgroup
  \vspace{0.5em}
}
\makeatother

\begin{document}

\maketitle


\newcommand{\teleopitTrackerPipelineFigure}{%
  \begin{figure}[t]
    \centering
    \includegraphics[width=\linewidth]{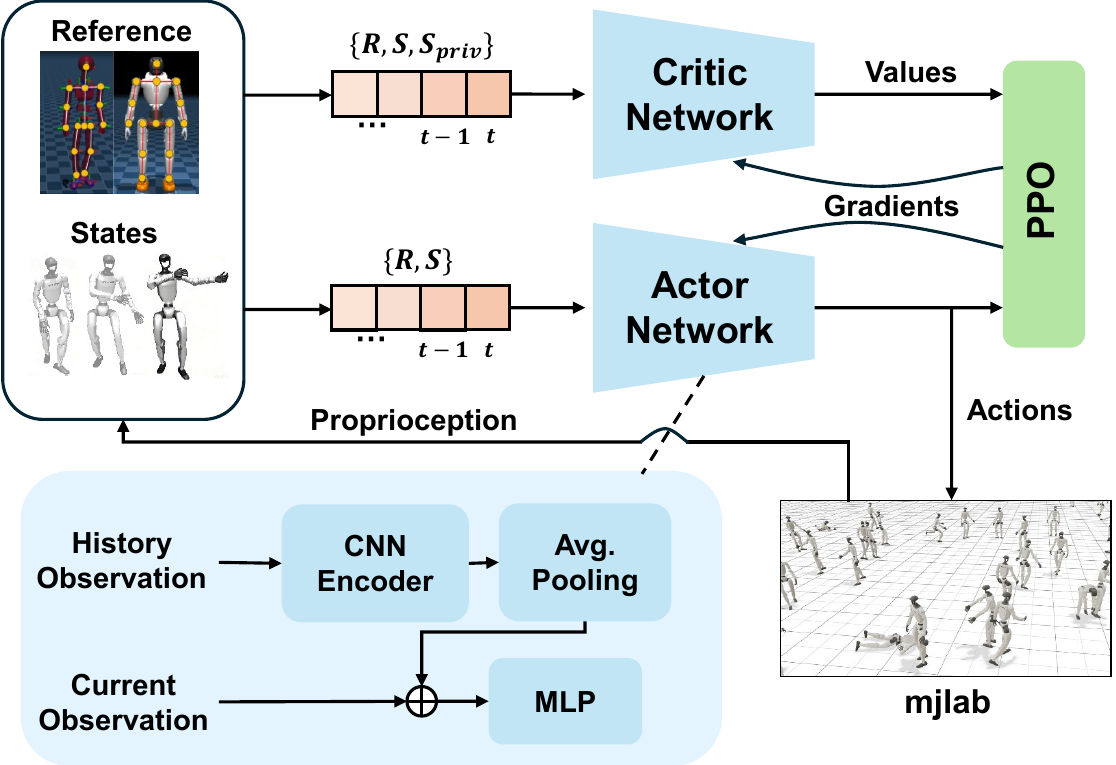}
    \caption{\textbf{Whole-body motion tracking pipeline.} The actor combines
      current proprioception and temporal history, while the training-only
      critic also uses privileged state. PPO optimizes predicted joint targets.}
    \label{fig:tracker-pipeline}
  \end{figure}
}

\newcommand{\teleopitRewindFigure}{%
  \begin{figure}[t]
    \centering
    \includegraphics[width=0.82\linewidth]{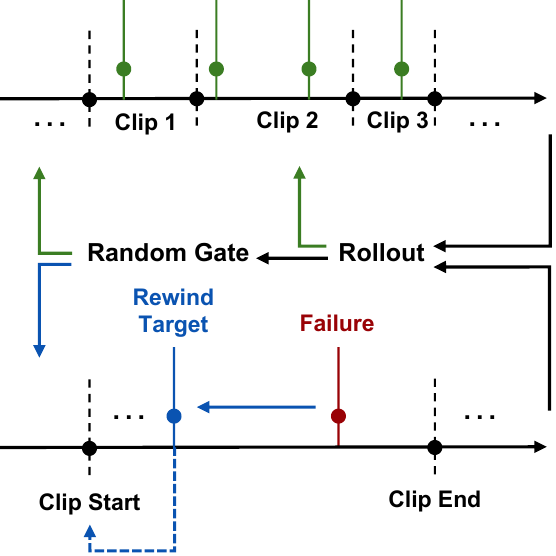}
    \caption{\textbf{Failure-aware rewind sampling.} Failed rollouts retain
      the clip and rewind before the failure, focusing subsequent updates on
      difficult transitions.}
    \label{fig:rewind-sampling}
  \end{figure}
}

\newcommand{\teleopitHandObjectivesFigure}{%
  \begin{figure}[t]
    \centering
    \includegraphics[width=\linewidth]{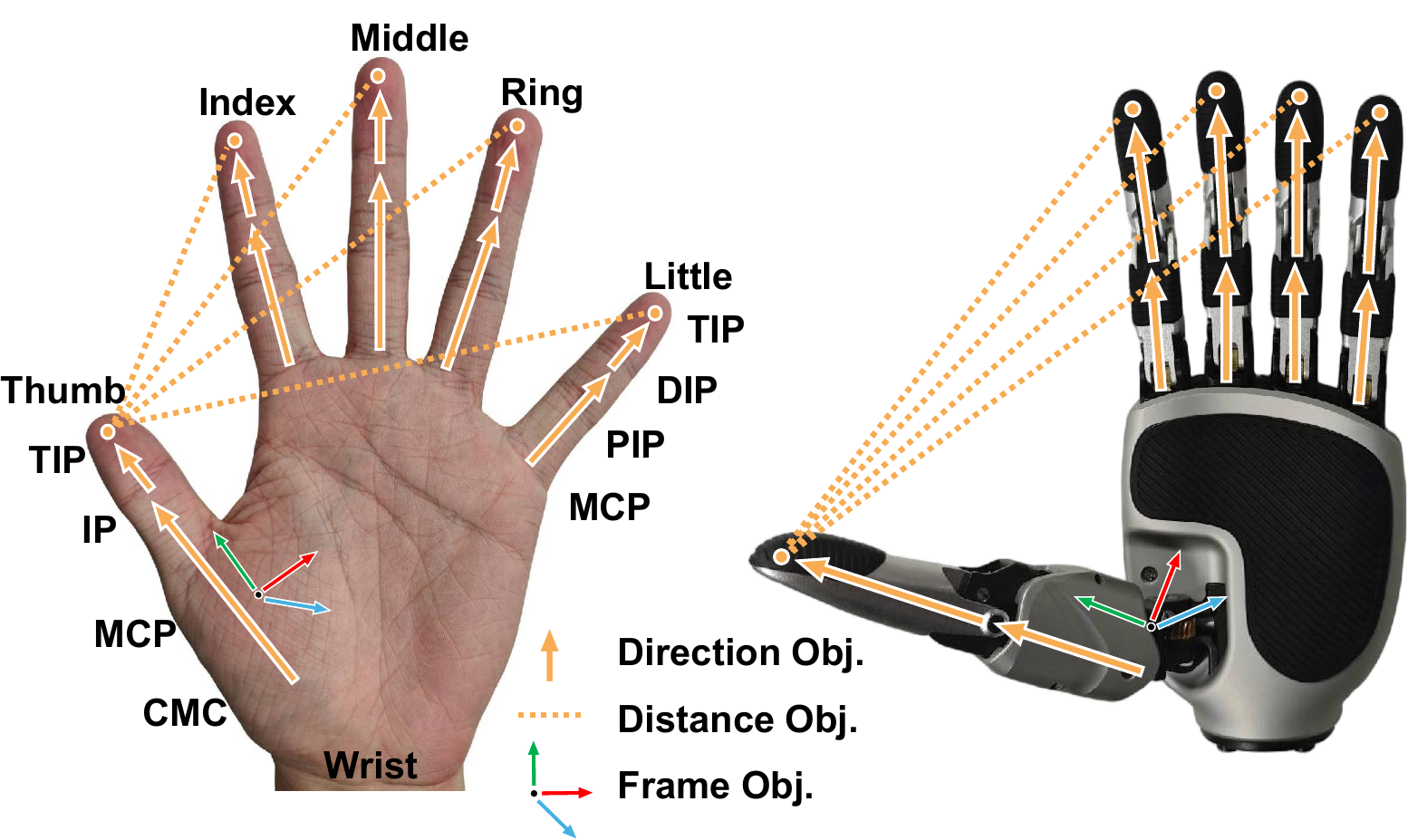}
    \caption{\textbf{Hand retargeting objectives.} Direction,
      fingertip-distance, and thumb-frame objectives map the human pose to a
      robot hand with different morphology.}
    \label{fig:dex-objective}
  \end{figure}
}

\newcommand{\teleopitDeploymentFigure}{%
  \begin{figure}[t]
    \centering
    \includegraphics[width=\linewidth]{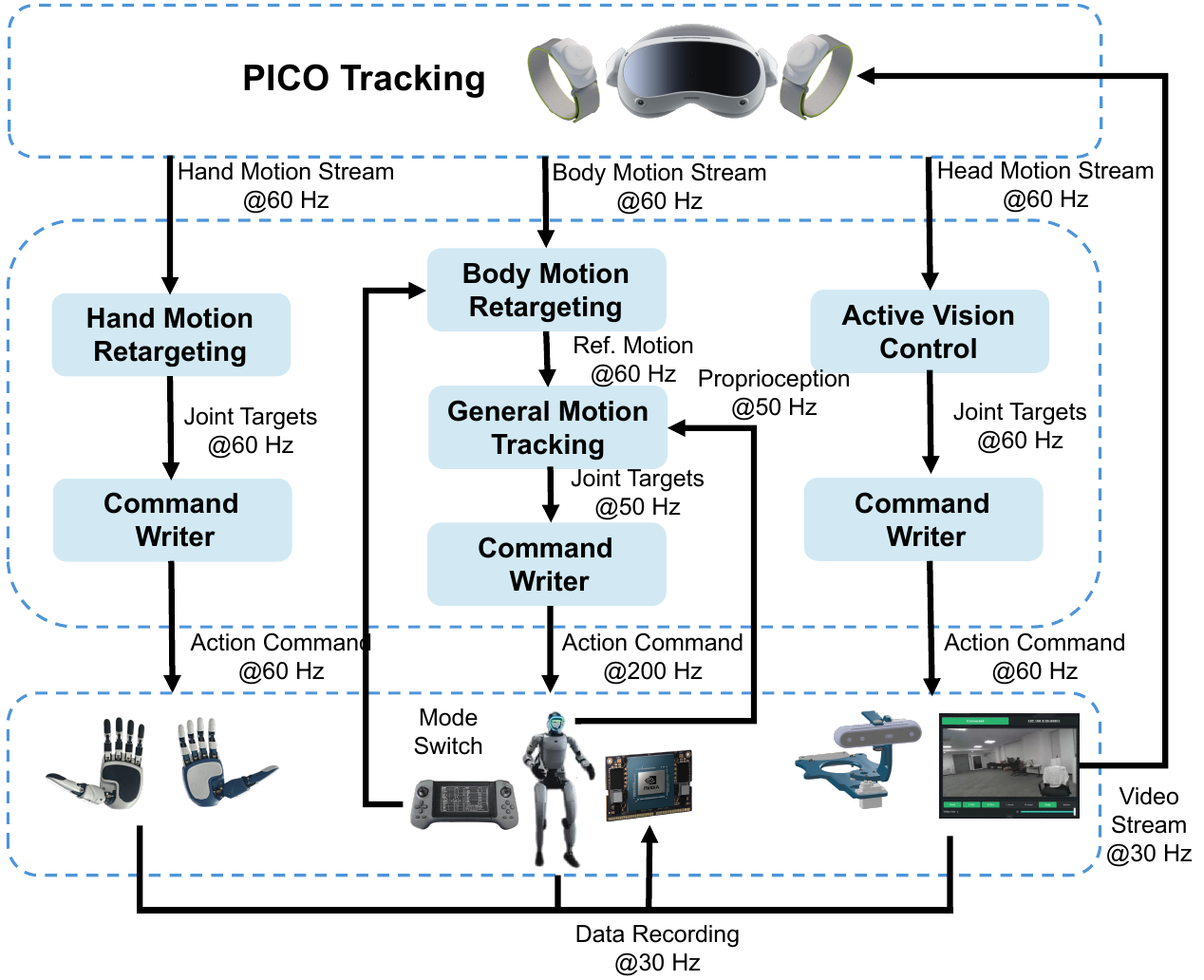}
    \caption{\textbf{Deployment architecture.} Asynchronous processes map
      PICO body, hand, and head streams to robot commands while returning video
      and recording synchronized data.}
    \label{fig:deployment-system}
  \end{figure}
}

\newcommand{\teleopitActiveVisionHardwareFigure}{%
  \begin{figure}[t]
    \centering
    \includegraphics[width=0.40\linewidth]{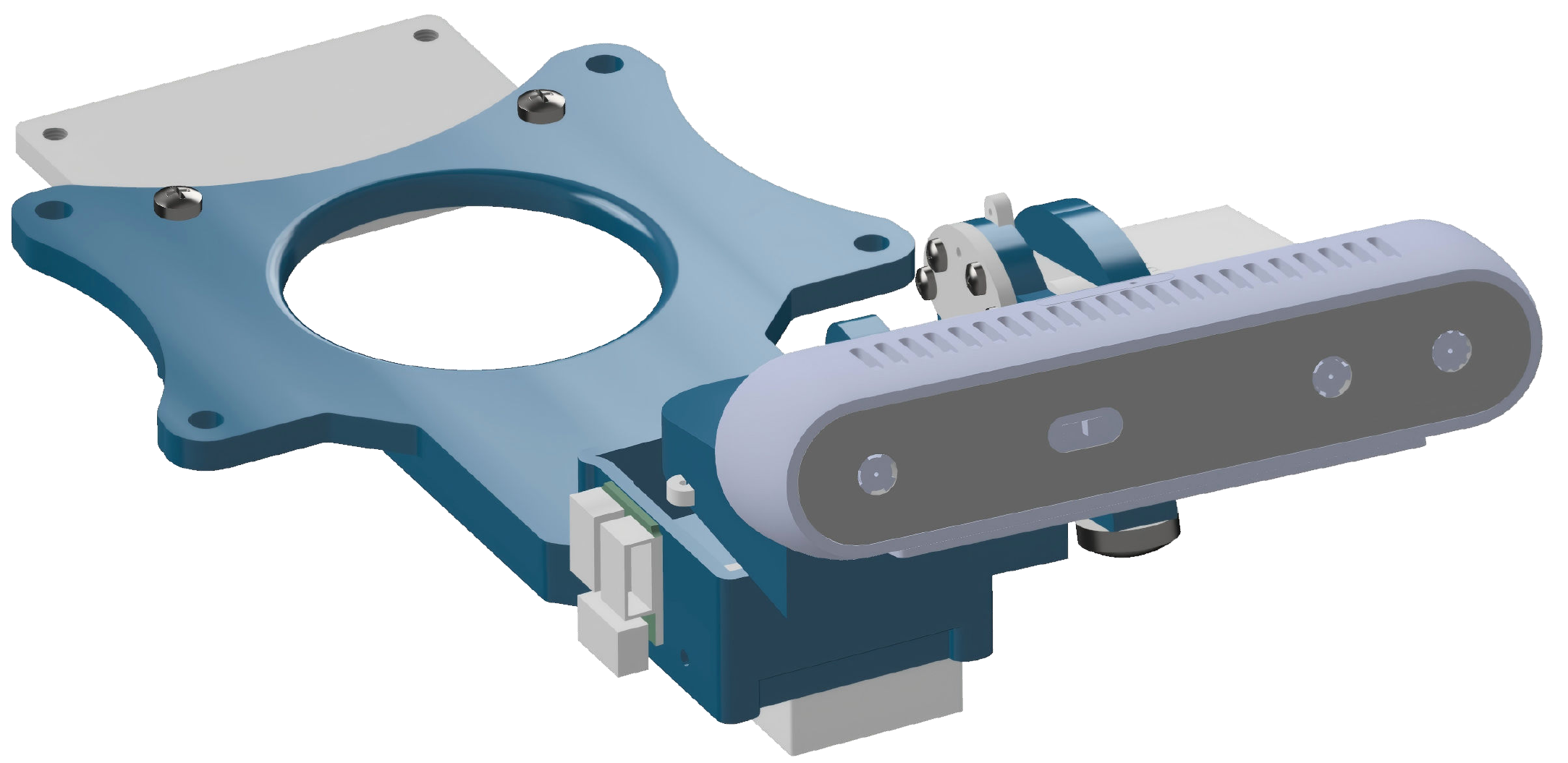}\hfill
    \includegraphics[width=0.40\linewidth]{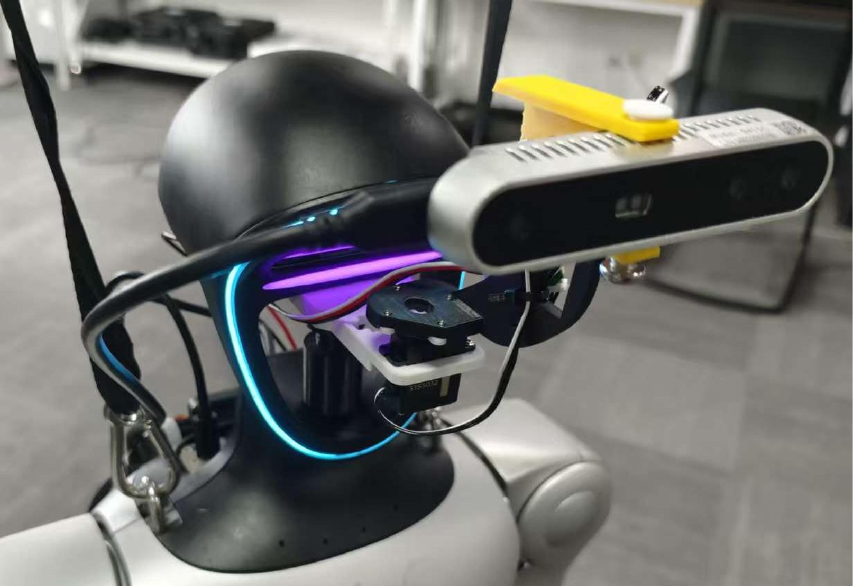}
    \caption{\textbf{Active vision module.} CAD and robot-mounted 2-DoF
      yaw--pitch camera.}
    \label{fig:active-vision-hardware}
  \end{figure}
}

\newcommand{\teleopitTrackerHardwareFigure}{%
  \begin{figure*}[t]
    \centering
    \includegraphics[width=\textwidth]{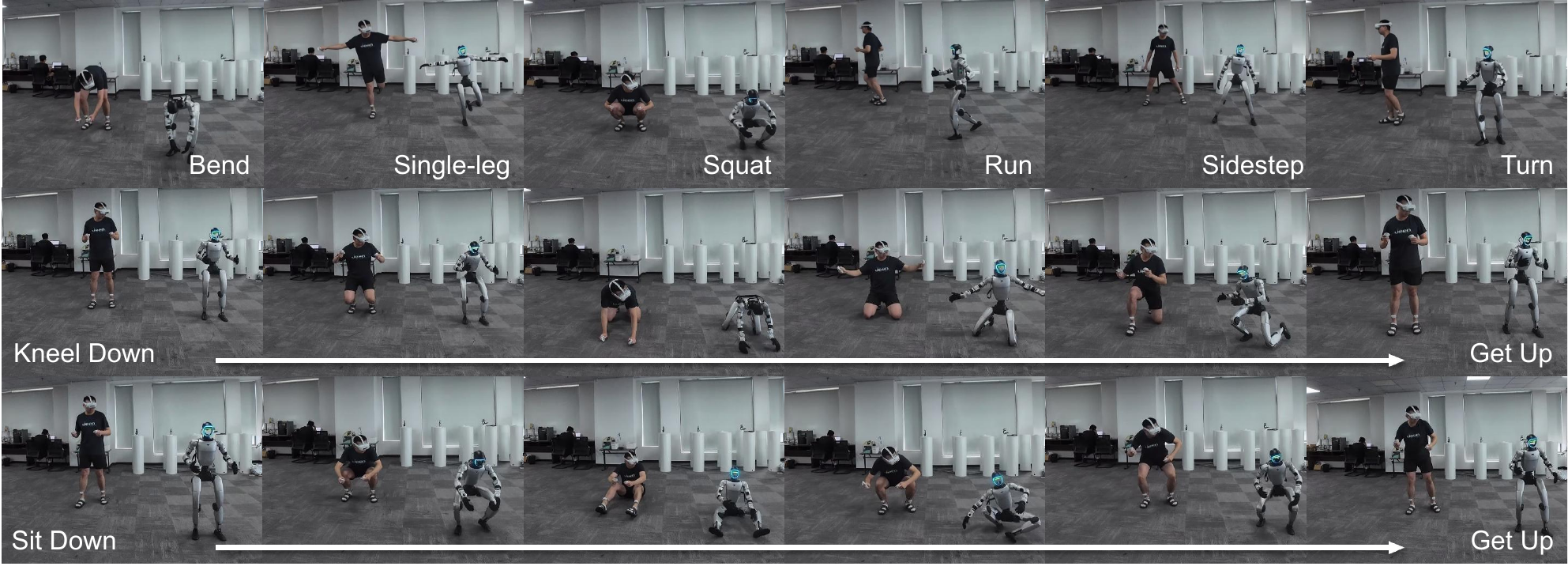}
    \caption{\textbf{Real-robot whole-body tracking.} Static motions and
      continuous kneel--stand and sit--stand transitions on hardware.}
    \label{fig:tracker-real}
  \end{figure*}
}

\newcommand{\teleopitHandPoseFigure}{%
  \begin{figure}[t]
    \centering
    \includegraphics[width=0.92\linewidth]{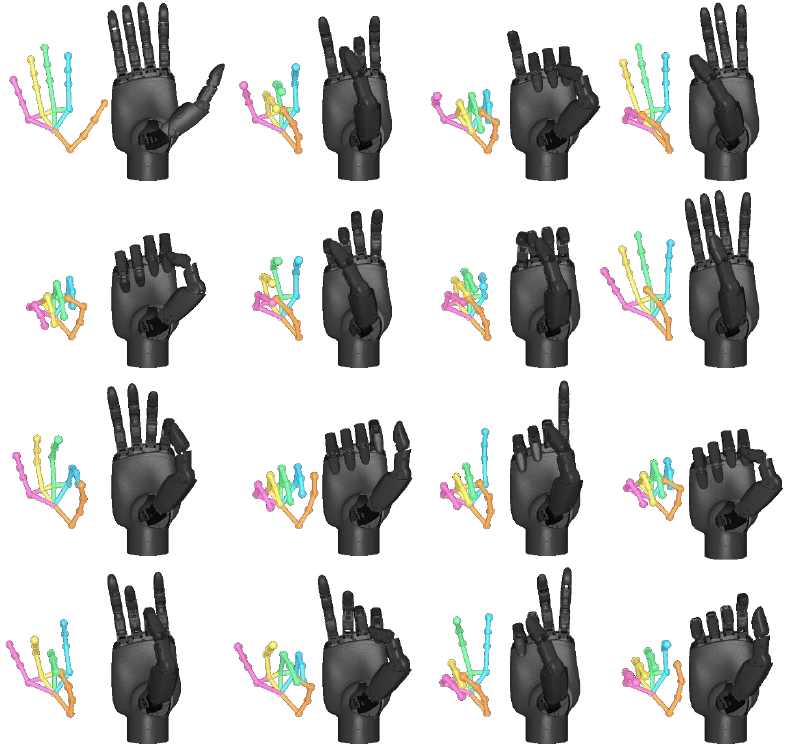}
    \caption{\textbf{Pose fidelity across gestures.} Human skeletons and
      retargeted poses show finger-direction, pinch, and thumb-opposition
      fidelity.}
    \label{fig:dex-multipose}
  \end{figure}
}

\newcommand{\teleopitHandTransferFigure}{%
  \begin{figure}[t]
    \centering
    \includegraphics[width=0.78\linewidth]{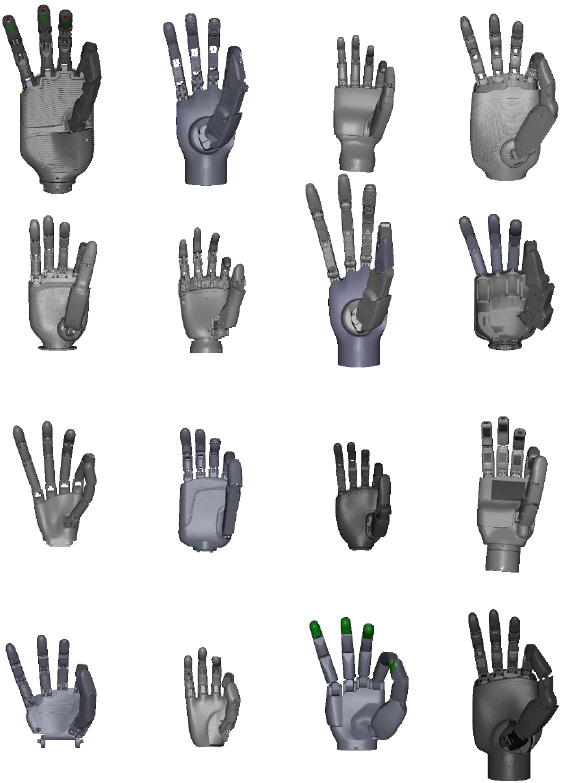}
    \caption{\textbf{Transfer across dexterous hands.} One objective handles
      different link lengths, finger counts, and thumb layouts; only semantic
      links and mechanical couplings change.}
    \label{fig:dex-multihand}
  \end{figure}
}

\newcommand{\teleopitHandAblationFigure}{%
  \begin{figure}[t]
    \centering
    \includegraphics[width=0.72\linewidth]{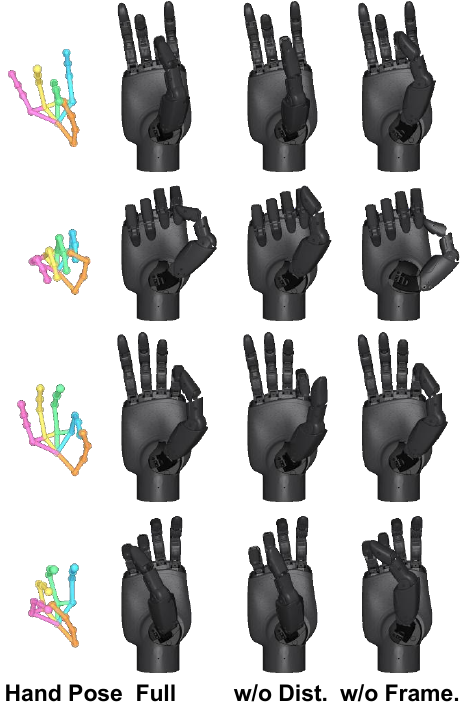}
    \caption{\textbf{Retargeting ablation.} The distance objective maintains
      pinch closure, and the frame objective preserves thumb opposition.}
    \label{fig:dex-ablation}
  \end{figure}
}

\newcommand{\teleopitIntegratedFigure}{%
  \begin{figure*}[t]
    \centering
    \includegraphics[width=\textwidth]{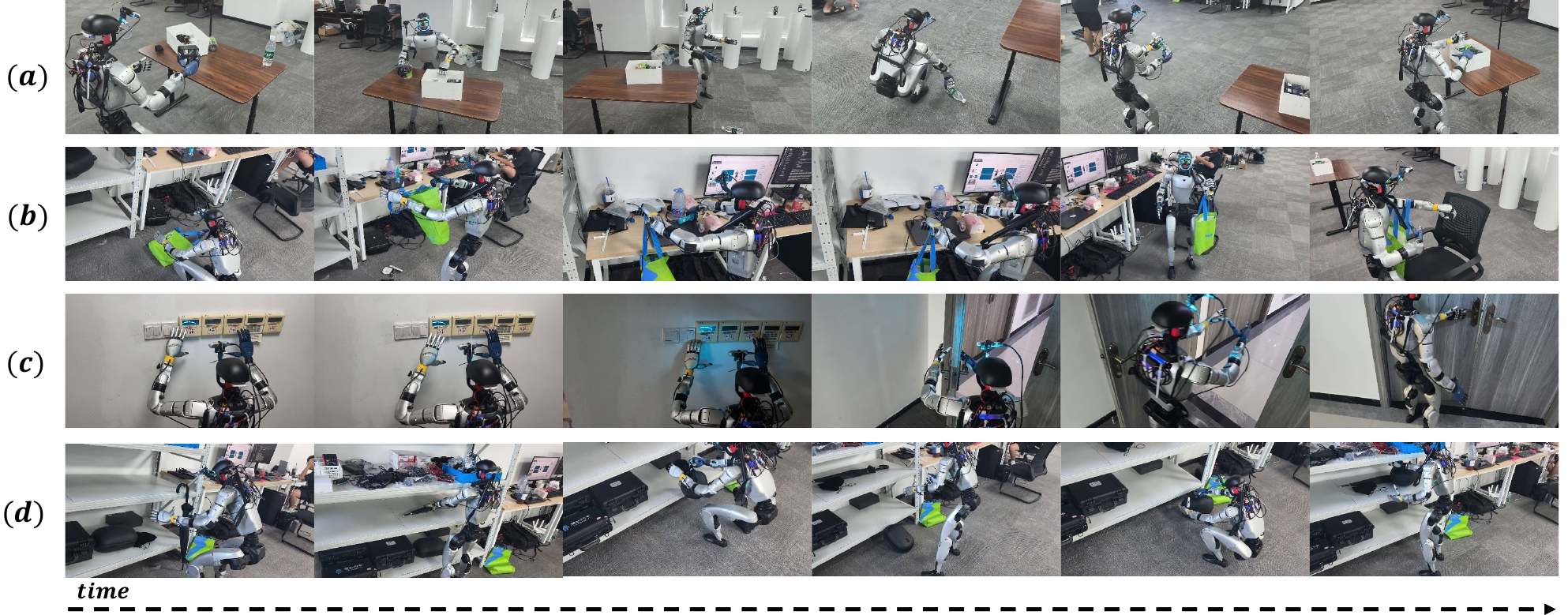}
    \caption{\textbf{Integrated loco-manipulation.} Body, hand, and vision
      control across four household tasks.}
    \label{fig:integrated-teleoperation}
  \end{figure*}
}

\section{Introduction}

Humanoid teleoperation provides a direct interface for collecting real-world
demonstrations for data-driven robot learning.
Vision-language-action models and robot foundation models increasingly benefit
from broad real-world robot datasets~\citep{brohan2023rt1roboticstransformerrealworld,%
brohan2023rt2visionlanguageactionmodelstransfer,%
embodimentcollaboration2025openxembodimentroboticlearning,%
khazatsky2025droidlargescaleinthewildrobot,%
kim2024openvlaopensourcevisionlanguageactionmodel,%
black2026pi0visionlanguageactionflowmodel,%
nvidia2025gr00tn1openfoundation}.
Unlike tabletop manipulation, humanoid demonstrations require balance,
locomotion, whole-body manipulation, continuous hand motion, and viewpoint
control to remain coordinated throughout a task.

Existing interfaces address only parts of this coordination problem.
VR systems can provide body, hand, and head signals without dedicated motion
capture equipment, but current humanoid systems commonly use only a subset of
these signals: some control the upper body and hands without locomotion, while
others control the whole body but reduce the hands to discrete gripper
commands~\citep{cheng2024opentelevisionteleoperationimmersiveactive,%
ze2025twist2scalableportableholistic}.
Systems that achieve continuous whole-body and dexterous hand control instead
rely on custom inertial suits and gloves~\citep{heng2026humdexhumanoiddexterousmanipulation}.
General XR frameworks simplify device integration but do not by themselves
provide a dynamically controlled humanoid
pipeline~\citep{zhao2025xrobotoolkitcrossplatformframeworkrobot}.
Moreover, dexterous hand retargeting must accommodate different link lengths,
joint layouts, and thumb structures; existing optimization methods require
hand-dependent geometric scaling, while learned mappings require training data
for the target hand~\citep{handa2019dexpilotvisionbasedteleoperation,%
qin2024anyteleopgeneralvisionbaseddexterous,%
yin2025geometricretargetingprincipledultrafast}.
These limitations leave no single workflow that combines VR sensing,
dynamically feasible whole-body control, continuous cross-hand retargeting,
viewpoint control, and recording.

We introduce \method{} (Figure~\ref{fig:system-teaser}), a full-embodiment
humanoid teleoperation system that uses VR as a unified source of operator
intent.
A body tracker maps the operator's motion to dynamically feasible humanoid
control, an optimization-based retargeter maps hand tracking to configurable
dexterous hands, and an active vision module maps head motion to viewpoint
control.
An asynchronous runtime connects these components with visual feedback and
recording, enabling \textbf{joint whole-body, hand, and viewpoint control}
while keeping each control stream responsive at its respective rate.

The whole-body tracker targets stable control from temporally correlated and
occasionally noisy VR references.
Its \textbf{history encoder} integrates recent reference and proprioceptive
observations so that the policy can infer motion context that is unavailable
from a single frame.
During training in mjlab~\citep{zakka2026mjlablightweightframeworkgpuaccelerated},
\textbf{failure-aware rewind sampling} returns failed rollouts to the difficult
segment that preceded termination, increasing exposure to transitions that
interrupt teleoperation.

The optimization-based hand retargeter addresses morphology differences,
building on vector-based dexterous
teleoperation~\citep{handa2019dexpilotvisionbasedteleoperation,%
qin2024anyteleopgeneralvisionbaseddexterous}.
Normalized finger directions remove bone-length scale from the primary
objective, while fingertip-distance and thumb-frame objectives encode pinch
closure and thumb opposition.
After semantic links are specified for a robot hand, the same objective weights,
activation thresholds, and solver settings map human motion to different hand
morphologies, providing \textbf{cross-embodiment transfer without hand-specific
tuning}.

We evaluate the motion tracker on held-out motion-capture and live VR
references, including comparisons with recent trackers and ablations of the
history encoder and rewind sampling.
Hand experiments examine cross-embodiment pose transfer and the effects of the
distance and frame objectives.
We further conduct real-robot teleoperation on tasks that coordinate whole-body
motion, continuous hand control, and viewpoint control.
Finally, we use 96 successful demonstrations collected with \method{} to
train ACT and GR00T N1.7. When deployed on the humanoid, the resulting policies
achieve task success rates of 90.0\% and 95.0\%, respectively, on a
bottle-placement task.

Our contributions are summarized as follows:
\begin{itemize}
    \item A VR-driven humanoid teleoperation system for collecting
    demonstrations with \textbf{joint whole-body, hand, and viewpoint
    control}.
    \item An optimization-based hand retargeter providing
    \textbf{cross-embodiment transfer without hand-specific tuning}.
    \item A motion tracker with a \textbf{history encoder} and
    \textbf{failure-aware rewind sampling} for robust live VR tracking.
\end{itemize}

\section{Related Work}

\subsection{Humanoid Teleoperation and Data Collection}

Humanoid teleoperation has progressed from kinematic motion transfer and
balance-aware inverse kinematics to learned controllers that track human motion
in real time~\citep{miller2004motioncaptureinertialhumanoid,%
montecillopuente2010realtimewholebodytransfer,%
he2024learninghumantohumanoidrealtimewholebody,%
he2024omnih2ouniversaldexteroushumantohumanoid,%
fu2024humanplushumanoidshadowingimitation,%
ze2025twistteleoperatedwholebodyimitation}.
Recent systems extend control beyond the body by combining VR, visual feedback,
or dexterous hands for loco-manipulation~\citep{seo2023deepimitationhumanoidloco,%
cheng2024opentelevisionteleoperationimmersiveactive,%
ben2025homiehumanoidlocomanipulationisomorphic,%
li2025cloneclosedloopwholebodyhumanoid}.
Robot-free interfaces offer higher collection throughput but translate human
demonstrations to the robot offline, while general XR frameworks provide device
and robot interfaces without a complete dynamically controlled humanoid
workflow~\citep{nai2026humanoidmanipulationinterface,%
wang2026bifrostumibridgingrobotfree,%
zhao2025xrobotoolkitcrossplatformframeworkrobot}.

TWIST2 and HumDex are the closest complete systems.
TWIST2 integrates PICO body sensing, full-body tracking, and active vision, but
controls its robot hands as discrete grippers~\citep{ze2025twist2scalableportableholistic}.
HumDex continuously controls a dexterous hand and the humanoid body using a
custom inertial suit and commercial gloves~\citep{heng2026humdexhumanoiddexterousmanipulation}.
Teleopit instead obtains body, hand, and head signals from VR and integrates
whole-body tracking, continuous dexterous hand control, and viewpoint control.

\subsection{Humanoid Motion Tracking}

Physics-based motion tracking uses reference-conditioned policies, motion
priors, and unified representations to reproduce diverse humanoid
motions~\citep{peng2018deepmimicexampleguided,%
peng2021ampadversarialmotionpriors,%
luo2023perpetualhumanoidcontrolrealtime,%
luo2024universalhumanoidmotionrepresentations,%
cheng2024expressivewholebodycontrolhumanoid,%
ji2025exbody2advancedexpressivehumanoid,%
he2025hoverversatileneuralwholebody}.
Recent trackers improve coverage through adaptive sampling, mixtures of
experts, temporal context, specialized policies, or larger training
corpora~\citep{chen2025gmtgeneralmotiontracking,%
zhang2025trackmotionsdisturbances,%
ma2026robustgeneralizedhumanoidmotion,%
ze2025twist2scalableportableholistic,%
li2026telegatewholebodyhumanoidteleoperation,%
luo2026sonicsupersizingmotiontracking,%
chen2026holomotion1technicalreport}.
Teleopit combines temporal history with failure-aware sampling to improve the
success rate of a single tracker on both motion-capture and live VR references.

\subsection{Dexterous Hand Retargeting}

Dexterous hand retargeting must map human motion across differences in link
length, joint layout, and actuation.
Geometric methods optimize fingertip positions, joint and Cartesian objectives,
or selected hand vectors, but typically depend on calibration or
morphology-specific scaling~\citep{hu2004calibratinghumanhand,%
meeker2018intuitivehandteleoperation,%
meattini2020combinedjointcartesian,%
meattini2023humanrobothandmappingreview,%
handa2019dexpilotvisionbasedteleoperation,%
qin2024anyteleopgeneralvisionbaseddexterous}.
Learned methods shift computation from online optimization to target-hand
training, paired samples, or guided data collection~\citep{sivakumar2022robotictelekinesis,%
yin2025geometricretargetingprincipledultrafast,%
heng2026humdexhumanoiddexterousmanipulation,%
wang2026anydexrtcalibrationfreedexterous}.

Task-oriented approaches additionally preserve contact, grasp function, or
dynamic feasibility for specific manipulation
settings~\citep{mandi2025dexmachinafunctionalretargetingbimanual,%
pan2026spiderscalablephysicsinformeddexterous,%
lee2026dextwistdexteroushandretargeting,%
qi2026genhandgeneralisedhuman}.
Teleopit instead uses a shared online optimization objective: normalized finger
directions handle morphology differences, while fingertip distance and thumb
frame objectives preserve pinch closure and thumb opposition.
New hands require semantic link mapping but no hand-specific objective or
solver hyperparameter tuning.

\section{Method}

\subsection{Overview}

Teleopit maps body, hand, and head signals from PICO to a humanoid robot in
real time. The input comprises a 24-joint body skeleton
$\mathbf{s}^{\textup{body}}$, 26 keypoints for each hand
$\mathbf{s}^{\textup{hand}}$, and a head pose $\mathbf{s}^{\textup{head}}$.
The outputs are body joint targets $\mathbf{a}^{\textup{body}}$, dexterous hand
commands $\mathbf{a}^{\textup{hand}}$, and a 2-DoF viewpoint command
$\mathbf{a}^{\textup{cam}}$.

The system contains a learned whole-body motion tracker, an optimization-based
hand retargeter, and a runtime that integrates sensing, control, visual
feedback, and recording. The following sections describe these components and
the motivation for their designs.

\subsection{Whole-Body Motion Tracking}
\label{sec:method-tracker}

\paragraph{Architecture.}
The tracker must convert a human reference into dynamically feasible robot
motion while remaining deployable with onboard state estimation.
As shown in Figure~\ref{fig:tracker-pipeline}, we train a single policy with
PPO~\citep{schulman2017proximalpolicyoptimizationalgorithms} in the
GPU-accelerated MuJoCo simulator
mjlab~\citep{zakka2026mjlablightweightframeworkgpuaccelerated}.
The actor receives robot proprioception and a reference frame and predicts a
$29$-dimensional action. We convert the action to joint targets as
\begin{equation}
  \mathbf{q}^{\textup{target}}
    = \textup{clip}(\mathbf{a}, -c, c)\odot\mathbf{s}
      + \mathbf{q}^{\textup{def}},
  \label{eq:action}
\end{equation}
where $\mathbf{s}$ is a per-joint action scale and
$\mathbf{q}^{\textup{def}}$ is the default pose. A PD controller tracks these
targets at $200$~Hz while the policy runs at $50$~Hz. Tracking is anchored at
the torso and covers 14 body links.

\teleopitTrackerPipelineFigure

\paragraph{History Encoder.}
A single observation frame does not reveal recent contact changes or body
momentum, which are important when VR references vary over time.
The History Encoder stacks the actor observation over the previous $H{=}10$
control steps and applies a one-dimensional temporal convolution followed by
global average pooling. The resulting latent is concatenated with the current
observation before the policy MLP, as illustrated in
Figure~\ref{fig:tracker-pipeline}. This supplies motion context without an
additional estimator or privileged teacher. The critic receives additional
simulation state during training, but these signals are not used by the actor.

\paragraph{Failure-Aware Rewind Sampling.}
Uniform sampling spends most updates on common motion while rarely revisiting
the difficult transitions that terminate teleoperation. Motivated by adaptive
sampling of difficult motion~\citep{chen2025gmtgeneralmotiontracking}, we use
the failure signal itself to identify hard segments.
Training initially samples clips in proportion to duration and resets the robot
to the sampled reference state. When a rollout terminates early, the strategy
in Figure~\ref{fig:rewind-sampling} retains the clip with high probability and
rewinds the reference time by a random offset. The policy therefore practices
the transition immediately before failure more frequently, without a separate
difficulty model or staged curriculum.

\teleopitRewindFigure

\paragraph{Observations and Rewards.}
The tracker combines deployable actor observations with privileged critic
signals and a dense whole-body tracking objective.

\paragraph{Observation Design}

The observation design separates deployable robot state from training-only
information. Table~\ref{tab:supp-tracker-obs} lists the per-step inputs. The
actor receives the reference motion, joint state, base angular velocity,
projected gravity, and previous action. These signals describe the target and
measured response without requiring global body poses. A ten-frame history
provides recent motion context.

The critic additionally receives global reference position, base linear
velocity, and the poses of the 14 tracked links, but the deployed actor never
sees them. Uniform corruption is applied to the channels marked in
Table~\ref{tab:supp-tracker-obs}; reference joint targets and the previous
action remain clean. This split keeps the actor inputs deployable while
enabling privileged value estimation during training.

\begin{table}[t]
  \centering
  \small
  \caption{Tracker observations. Noise is the half-width of uniform training
    corruption. A and C denote actor and critic inputs; the lower block is
    critic-only.}
  \label{tab:supp-tracker-obs}
  \begin{tabular}{@{}llcc@{}}
    \toprule
    \textbf{Term} & \textbf{Dim} & \textbf{Noise} & \textbf{A\,/\,C} \\
    \midrule
    Reference joint position & $29$ & -- & A, C \\
    Reference joint velocity & $29$ & -- & A, C \\
    Reference anchor orientation (6D) & $6$ & $0.05$ & A, C \\
    Reference anchor lin. velocity & $3$ & -- & A, C \\
    Reference anchor ang. velocity & $3$ & -- & A, C \\
    Reference projected gravity & $3$ & -- & A, C \\
    Reference anchor height & $1$ & -- & A, C \\
    Joint position (rel. to default) & $29$ & $0.01$ & A, C \\
    Joint velocity & $29$ & $0.5$ & A, C \\
    Base angular velocity & $3$ & $0.2$ & A, C \\
    Projected gravity & $3$ & $0.05$ & A, C \\
    Previous action & $29$ & -- & A, C \\
    \midrule
    Reference anchor position & $3$ & -- & C \\
    Base linear velocity & $3$ & -- & C \\
    Tracked-body position ($14$) & $42$ & -- & C \\
    Tracked-body orientation ($14$) & $84$ & -- & C \\
    \bottomrule
  \end{tabular}
\end{table}

\paragraph{Reward Design}

The reward combines global tracking, articulated pose tracking, and
deployment-oriented regularization. In
Table~\ref{tab:supp-tracker-reward}, anchor terms preserve the torso trajectory,
body and joint terms constrain the full pose and dynamics, and the wrist-palm
point directly targets hand placement for downstream dexterous control.

Gaussian scales determine the informative error range of each tracking term.
The remaining terms discourage rapid commands, joint-limit violations,
self-collision, and high ankle acceleration, while rewarding survival.
Together, the terms encode both pose tracking and motion regularity rather
than a single joint-space error.

\begin{table}[t]
  \centering
  \small
  \setlength{\tabcolsep}{4pt}
  \caption{Tracker reward. Tracking terms use a Gaussian kernel over squared
    error $e$; regularization terms use the listed penalties.}
  \label{tab:supp-tracker-reward}
  \begin{tabular}{@{}llr@{}}
    \toprule
    \textbf{Term} & \textbf{Expression} & \textbf{Weight} \\
    \midrule
    \multicolumn{3}{@{}l}{\textit{Tracking}\quad($r=\exp(-e/\sigma^2)$)} \\
    Anchor position       & $\sigma=0.3$   & $0.5$ \\
    Anchor orientation    & $\sigma=0.4$   & $0.5$ \\
    Anchor linear vel.    & $\sigma=1.0$   & $1.0$ \\
    Anchor angular vel.   & $\sigma=3.0$   & $1.0$ \\
    Body position         & $\sigma=0.3$   & $1.0$ \\
    Body orientation      & $\sigma=0.4$   & $1.0$ \\
    Body linear vel.      & $\sigma=1.0$   & $1.0$ \\
    Body angular vel.     & $\sigma=3.14$  & $1.0$ \\
    Joint position        & $\sigma=0.5$   & $1.0$ \\
    Joint velocity        & $\sigma=3.0$   & $0.5$ \\
    Wrist palm point      & $\sigma=0.12$  & $1.0$ \\
    \midrule
    \multicolumn{3}{@{}l}{\textit{Regularization}} \\
    Survival              & $\mathbf{1}$ (alive) & $3.0$ \\
    Action rate           & $\lVert a_t-a_{t-1}\rVert^2$ & $-0.5$ \\
    Joint-limit violation & $\sum_j\max(0,q_j-q_j^{\lim})$ & $-10.0$ \\
    Self-collision        & $\mathbf{1}[f_{\text{self}}>1\text{N}]$ & $-0.1$ \\
    Ankle joint accel.    & $\lVert\ddot q_{\text{ankle}}\rVert^2$ & $-2.5\times10^{-6}$ \\
    \bottomrule
  \end{tabular}
\end{table}

\paragraph{Domain Randomization Design}

The domain randomization mechanisms cover persistent physical variation,
sensing noise, and transient disturbances. At each environment reset, we
randomize friction, torso center of mass, link inertia, and joint offsets.
Per-step observation noise perturbs selected measurements, while periodic
base-velocity perturbations expose the policy to transient disturbances.
Table~\ref{tab:supp-tracker-dr} lists the range of each perturbation.

\paragraph{Reference Sampling and Termination}

Reference-state initialization resets each environment to a sampled pose with
the perturbations in Table~\ref{tab:supp-tracker-dr}, and clips are sampled in
proportion to their valid duration. When a rollout terminates early, rewind
sampling retains the same clip with probability $0.8$ and moves the reference
to an earlier state near the failed transition.

An episode terminates when the anchor height error exceeds $0.25$~m, the anchor
orientation error exceeds $1.0$~rad, or the height error of an ankle or wrist
exceeds $0.25$~m. These criteria supply the failure event used by rewind and
prevent severely diverged states from dominating subsequent updates. Thus,
domain randomization broadens the physical and sensory conditions encountered
during training, while reference-state initialization and rewind determine
where training effort is spent along each motion clip.

\begin{table}[t]
  \centering
  \small
  \caption{Domain randomization, observation noise, and reference-state
    initialization for the tracker. $\pm x$ denotes a uniform draw over
    $[-x, x]$.}
  \label{tab:supp-tracker-dr}
  \begin{tabular}{@{}ll@{}}
    \toprule
    \textbf{Parameter} & \textbf{Range} \\
    \midrule
    \multicolumn{2}{@{}l}{\textit{Dynamics (startup)}} \\
    Ground friction & $[0.3, 1.6]$ \\
    Torso CoM offset $(x,y,z)$ [m] & $(\pm0.025, \pm0.05, \pm0.05)$ \\
    Link pseudo-inertia scale $\alpha$ & $[-0.1, 0.45]$ \\
    Default joint-position offset [rad] & $\pm0.01$ \\
    \midrule
    \multicolumn{2}{@{}l}{\textit{Observation noise (uniform)}} \\
    Joint position [rad] & $\pm0.01$ \\
    Joint velocity [rad/s] & $\pm0.5$ \\
    Base angular velocity [rad/s] & $\pm0.2$ \\
    Projected gravity & $\pm0.05$ \\
    Anchor orientation & $\pm0.05$ \\
    \midrule
    \multicolumn{2}{@{}l}{\textit{Random push (every $4$--$6$~s)}} \\
    Base linear velocity $(x,y,z)$ [m/s] & $(\pm0.5, \pm0.5, \pm0.2)$ \\
    Base angular velocity (r,p,y) [rad/s] & $(\pm0.52, \pm0.52, \pm0.78)$ \\
    \midrule
    \multicolumn{2}{@{}l}{\textit{Reference-state init.}} \\
    Root position $(x,y,z)$ [m] & $(\pm0.05, \pm0.05, \pm0.01)$ \\
    Root orientation (r,p,y) [rad] & $(\pm0.1, \pm0.1, \pm0.2)$ \\
    Root velocity & same as push \\
    Joint position [rad] & $\pm0.1$ \\
    \bottomrule
  \end{tabular}
\end{table}

\subsection{Dexterous Hand Retargeting}
\label{sec:method-hand}

\paragraph{Formulation and Notation.}
Human and robot hands differ in bone lengths, joint layouts, and thumb
articulation. We therefore optimize geometric relations that can be defined on
both embodiments, as shown in Figure~\ref{fig:dex-objective}.
Let $\mathbf{q}$ denote the robot hand configuration,
$\mathbf{v}_i(\mathbf{q})$ a robot finger segment, and
$\hat{\mathbf{d}}_i$ the corresponding unit human segment direction.
For fingertip pair $k$, $\rho_k(\mathbf{q})$ and $\hat{\rho}_k$ denote robot
and target distances. Robot thumb-frame axes are
$\mathbf{e}_p(\mathbf{q}),\mathbf{e}_s(\mathbf{q})$, with human references
$\hat{\mathbf{e}}_p,\hat{\mathbf{e}}_s$.
The retargeted pose is
\begin{equation}
  \mathbf{q}^\star = \arg\min_{\mathbf{q}}
    \mathcal{L}_{\textup{dir}} + \mathcal{L}_{\textup{dist}}
    + \mathcal{L}_{\textup{frame}}
    + \lambda\lVert\bar{\mathbf{q}}-\bar{\mathbf{q}}_{\textup{prev}}\rVert^2,
  \label{eq:hand-obj}
\end{equation}
subject to the robot joint limits. The final term smooths the expanded joint
configuration $\bar{\mathbf{q}}$ over time.

\teleopitHandObjectivesFigure

\paragraph{Direction Objective.}
The orange arrows in Figure~\ref{fig:dex-objective} align corresponding segments
through
\begin{equation}
  \mathcal{L}_{\textup{dir}}
    = \sum_i w_i\left(1 -
      \frac{\mathbf{v}_i(\mathbf{q})}{\lVert\mathbf{v}_i(\mathbf{q})\rVert}
      \cdot\hat{\mathbf{d}}_i\right).
  \label{eq:hand-dir}
\end{equation}
Normalizing both segments removes bone length from the comparison. Unlike
scaled keyvector objectives~\citep{handa2019dexpilotvisionbasedteleoperation,%
qin2024anyteleopgeneralvisionbaseddexterous,%
seed2025bytedexterdexterousteleoperation}, this objective does not require a
human-to-robot scale for each hand.

\paragraph{Distance Objective.}
Direction alignment does not guarantee fingertip closure. The dotted lines in
Figure~\ref{fig:dex-objective} identify thumb-to-fingertip pairs, for which we
use
\begin{equation}
  \mathcal{L}_{\textup{dist}}
    = \sum_k w_k a_k
      \max\!\left(0,\rho_k(\mathbf{q})-\hat{\rho}_k\right)^2.
  \label{eq:hand-dist}
\end{equation}
The activation $a_k$ increases smoothly when the human fingertips approach a
4~cm contact threshold. The one-sided objective closes a commanded pinch
without pushing fingertips apart during free hand motion.

\paragraph{Frame Objective.}
A single thumb segment does not determine opposition about its axis. The axes
at the thumb base in Figure~\ref{fig:dex-objective} define a local frame from
the thumb CMC, thumb MCP, and index MCP. Gram--Schmidt orthonormalization
produces the two reference axes, which are aligned by
\begin{equation}
  \mathcal{L}_{\textup{frame}}
    = w_p\left(1-\mathbf{e}_p(\mathbf{q})\cdot\hat{\mathbf{e}}_p\right)
    + w_s\left(1-\mathbf{e}_s(\mathbf{q})\cdot\hat{\mathbf{e}}_s\right).
  \label{eq:hand-frame}
\end{equation}
This objective resolves thumb roll and supports both precision and power
grasp poses.

\paragraph{Optimization and Hand Configuration.}
The 21 human keypoints are expressed in a wrist frame derived from the wrist,
index MCP, and middle MCP. We solve Equation~\eqref{eq:hand-obj} for each frame
with SLSQP, analytic kinematic gradients, and the previous solution as the
initial state. The objective terms reference semantic link names, so a new
hand only specifies the corresponding robot links.
For mechanically coupled hands, independent coordinates $\mathbf{q}$ are
expanded through a coupling map
$\bar{\mathbf{q}}=\phi(\mathbf{q})$, and gradients propagate through $\phi$.
All configured hands share the numerical objective weights, activation
thresholds, and solver settings; Section~\ref{sec:exp-hand} evaluates this
cross-embodiment transfer.

\subsection{System Integration}
\label{sec:method-system}

Figure~\ref{fig:deployment-system} summarizes the runtime that connects the
motion tracker, hand retargeter, and active vision module. It consists of a
bidirectional XR bridge, a 2-DoF camera mechanism, and asynchronous control and
recording processes.

\teleopitDeploymentFigure

\paragraph{Bidirectional XR Bridge.}
The bridge streams body, hand, head, and controller signals to the host and
returns the robot camera view to the headset.

\begin{figure}[t]
  \centering
  \includegraphics[width=0.72\linewidth]{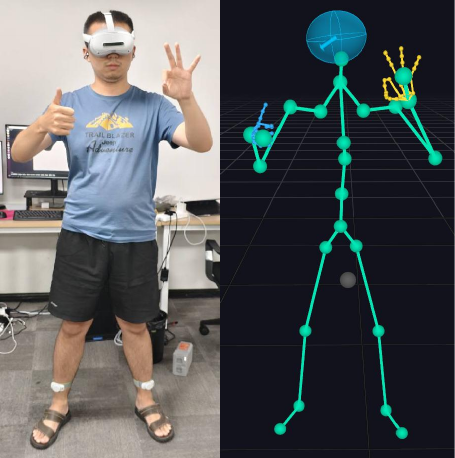}
  \caption{\textbf{PICO motion interface.} Host-side visualization of the body
    skeleton, head pose, and articulated hand keypoints reconstructed from the
    headset and ankle-tracker signals.}
  \label{fig:supp-pico-bridge}
\end{figure}

The XR bridge converts PICO outputs into a common host representation consumed
by the motion tracker, hand retargeter, and viewpoint
controller. Figure~\ref{fig:supp-pico-bridge} visualizes the reconstructed body
skeleton, head pose, and articulated hand keypoints. Keeping these signals in a
shared timestamped representation allows the downstream controllers and
recorder to associate commands that originate from the same operator motion.

The headset sends the body skeleton, hand keypoints, head pose, and controller
inputs to the host through timestamped TCP messages. UDP discovery removes
manual address configuration, and WebRTC returns the robot camera stream to
the headset. Latest-only queues favor fresh commands over intermediate poses
and prevent transient delays from accumulating into a processing backlog.
Together with the freshness checks described below, the sensing,
control, video, and recording processes run asynchronously.

\paragraph{Active Vision Module.}
The module maps head orientation relative to the torso to yaw and pitch
commands after dead-zone filtering and exponential smoothing. Relative head
motion prevents torso rotation from changing the commanded viewpoint. Two
serial-bus servos actuate the camera, and the returned video closes the
viewpoint control loop. The mechanism in
Figure~\ref{fig:active-vision-hardware} costs approximately CNY~500 excluding
the camera.

\teleopitActiveVisionHardwareFigure

\paragraph{Asynchronous Execution.}
Separate processes run PICO capture, hand retargeting, viewpoint control, body
policy inference, robot communication, video, and recording at their native
rates. PICO signals, hand retargeting, and viewpoint control run at 60~Hz; the
body policy runs at 50~Hz; low-level body control runs at 200~Hz; and recording
runs at 30~Hz. Latest-only communication keeps control responsive, while
freshness checks hold the last safe command when a stream becomes stale.

\section{Experiments}

We organize the evaluation around four questions. We first evaluate whether the
whole-body motion tracker balances tracking fidelity and robustness, including
comparisons, ablations, and deployment on the humanoid. We then examine whether
the hand retargeter preserves dexterous poses across different hand
morphologies and isolate the effect of its secondary objectives. Next, we
evaluate the integrated system through closed-loop viewpoint control and
loco-manipulation demonstrations on the humanoid. Finally, we examine whether
demonstrations recorded with \method{} can train policies for autonomous task
execution on the humanoid.

\subsection{Whole-Body Motion Tracking}
\label{sec:exp-tracker}

The tracker evaluation combines a controlled simulation benchmark with
real-world deployment evidence. The benchmark measures tracking accuracy and
rollout-level robustness on held-out mocap and live PICO references, while the
hardware sequences test whether one policy covers the motions required during
teleoperation.

\subsubsection{Training Setup}
\label{sec:exp-tracker-impl}

\paragraph{Datasets.}
\label{sec:exp-datasets}

We train the tracker on heterogeneous retargeted whole-body motion from three
public mocap sources:
BONES-SEED~\citep{bones2026seed},
TWIST2~\citep{ze2025twist2scalableportableholistic}, and
LAFAN1~\citep{harvey2020robust}, together with PICO teleoperation recordings
captured by our system. Evaluation uses separate mocap and PICO validation
subsets to distinguish clean retargeted references from the noisier references
encountered during deployment.

The training set combines retargeted public motion with PICO teleoperation
recordings to cover diverse motion and the deployment reference distribution.
Table~\ref{tab:supp-datasets} reports the source composition. The validation
data are disjoint from training and contain a retargeted mocap subset and a
recorded PICO subset.

The listed validation sources contain 238 mocap clips and 5 PICO recordings.
For evaluation, we retain windows that span the complete 10-second horizon,
which yields 181 mocap windows and 67 PICO windows. This windowing prevents a
short source segment from being counted as a successful full-length rollout.

\begin{table}[t]
  \centering
  \small
  \setlength{\tabcolsep}{6pt}
  \caption{Training and validation sets for the \method{} motion tracker.
    Duration is reported as \texttt{h:mm:ss}.}
  \label{tab:supp-datasets}
  \begin{tabular}{@{}lrr@{}}
    \toprule
    \textbf{Source} & \textbf{Clips} & \textbf{Duration} \\
    \midrule
    \multicolumn{3}{@{}l}{\textit{Training}} \\
    Subset of BONES-SEED & 63,237 & 127:56:57 \\
    Subset of TWIST2 & 31,071 &  88:14:44 \\
    Subset of LAFAN1 &     57 &   3:22:58 \\
    PICO recordings  &      9 &   0:31:19 \\
    \midrule
    \multicolumn{3}{@{}l}{\textit{Validation}} \\
    Subset of BONES-SEED & 238 & 0:53:44 \\
    PICO recordings &   5 & 0:16:20 \\
    \bottomrule
  \end{tabular}
\end{table}

\paragraph{Simulation and training.}
We train the tracker with PPO in the GPU-accelerated mjlab simulator. The policy
combines current-frame features with a temporal encoding of the previous
$10$ observations, while the critic additionally uses privileged simulation
state only during training.

We train the tracker with mjlab on 8 NVIDIA A800 GPUs for approximately
50 hours. The actor and critic use the same architecture: an MLP processes the
current observation, a one-dimensional temporal convolution encodes the
10-frame history, and the resulting latent is concatenated with the
current-frame features before the policy or value MLP.
The critic additionally receives the privileged signals in
Table~\ref{tab:supp-tracker-obs}; these signals are never used by the deployed
actor.

Tracking is anchored at the \texttt{torso\_link} and covers 14 body links:
the pelvis; both hip-roll, knee, and ankle-roll links; the torso; and both
shoulder-roll, elbow, and wrist-yaw links. The policy outputs 29 joint-target
offsets from the default pose. Table~\ref{tab:supp-tracker-hparams} reports the
complete simulation, network, and PPO configuration.

\begin{table}[t]
  \centering
  \small
  \setlength{\tabcolsep}{4pt}
  \caption{Simulation, network architecture, and PPO hyperparameters of the
    \method{} motion tracker.}
  \label{tab:supp-tracker-hparams}
  \begin{tabular}{@{}ll@{}}
    \toprule
    \textbf{Setting} & \textbf{Value} \\
    \midrule
    \multicolumn{2}{@{}l}{\textit{Simulation \& control}} \\
    Simulator & mjlab (GPU MuJoCo) \\
    Robot & Unitree G1, $29$ DoF \\
    Physics timestep & $5$~ms ($200$~Hz) \\
    Control decimation & $4$ \\
    Policy / PD frequency & $50$ / $200$~Hz \\
    Episode length & $10$~s \\
    Training hardware & 8 NVIDIA A800 GPUs \\
    Parallel environments & $8192$/GPU ($65\,536$ total) \\
    Training wall time & approximately 50 hours \\
    \midrule
    \multicolumn{2}{@{}l}{\textit{Architecture}} \\
    Actor/critic MLP dims & $[2048, 1024, 512, 256, 128]$ \\
    Activation & ELU \\
    History encoder & Conv1d $[256, 128, 64]$ \\
    History enc.\ kernel / pool & $3$ / global average \\
    Observation history length $H$ & $10$ \\
    Observation normalization & empirical (running) \\
    Policy distribution & Gaussian, init.\ std $1.0$ \\
    \midrule
    \multicolumn{2}{@{}l}{\textit{Training (PPO)}} \\
    Steps per environment & $24$ \\
    Learning epochs & $5$ \\
    Mini-batches & $4$ \\
    Learning rate & $5\times10^{-4}$ (adaptive) \\
    Desired KL & $0.01$ \\
    Discount $\gamma$ & $0.99$ \\
    GAE $\lambda$ & $0.95$ \\
    Clip range & $0.2$ \\
    Entropy coefficient & $0.005$ \\
    Value-loss coefficient & $1.0$ \\
    Max gradient norm & $1.0$ \\
    Iterations & $40\,000$ \\
    \bottomrule
  \end{tabular}
\end{table}

\paragraph{Domain randomization and reference sampling.}
Training combines domain randomization, reference-state initialization,
duration-proportional clip sampling, and failure-aware rewind sampling.
Section~\ref{sec:method-tracker} and
Table~\ref{tab:supp-tracker-dr} give the complete randomization ranges,
reference sampling procedure, and termination criteria.

\subsubsection{Evaluation Setup}

The benchmark contains 181 mocap and 67 PICO windows, each spanning 10 seconds,
and every method performs one deterministic rollout per window. We compare
against three recent humanoid whole-body trackers:
SONIC~\citep{luo2026sonicsupersizingmotiontracking},
HoloMotion~\citep{chen2026holomotion1technicalreport}, and
TWIST2~\citep{ze2025twist2scalableportableholistic}. Because their released
trackers differ in training data and budgets, we recompute success rate for all
methods with the same \method{} termination conditions. We report success rate
over all rollouts and four tracking errors over successful rollouts: MPJPE,
root-position error, root-orientation error, and root-velocity error. Results
are separated by validation subset in Table~\ref{tab:tracker-results}. The complete evaluation protocol and metric definitions follow.

Each comparison method performs one deterministic 10-second rollout on every
validation window. We report the mocap and PICO subsets separately. The
comparison includes SONIC, HoloMotion, and TWIST2. Their released trackers use
different training data, reward functions, training budgets, and termination
conditions. At evaluation time, success rate is recomputed for every method
using the same \method{} termination criteria given in the preceding section.

Success rate is computed over every evaluation rollout. Continuous tracking
errors are averaged only over rollouts that reach the full 10-second timeout.
These errors therefore measure tracking fidelity conditional on completing
the full horizon, without mixing early-terminated trajectories of different
lengths.

\subsubsection{Metric Definitions}

Consider a rollout of $T$ control steps that tracks a reference with $J{=}14$
links. Let $\mathbf p_{t,j}\in\mathbb{R}^3$ denote the position of link $j$ at
step $t$. The vectors $\mathbf p^{\textup{root}}_{t}$ and
$\mathbf v^{\textup{root}}_{t}$ denote the anchor position and body-frame
linear velocity, and $\mathbf q^{\textup{root}}_{t}$ is its unit orientation
quaternion. A hat denotes the corresponding reference quantity. A tilde
denotes a quantity expressed in the root heading frame, so global translation
and heading are factored out.

\paragraph{Success Rate.}
Given $N$ evaluation rollouts, success rate (SR) is the fraction that reach the
full timeout without early termination:
\begin{equation}
  \textup{SR}=\frac{1}{N}\sum_{n=1}^{N}
    \mathbf{1}\!\left[\text{rollout }n\text{ reaches timeout}\right].
  \label{eq:supp-metric-sr}
\end{equation}

\paragraph{Mean Per-Joint Position Error.}
MPJPE measures pose fidelity after aligning the robot and reference in the root
heading frame:
\begin{equation}
  \textup{MPJPE}=\frac{1}{TJ}\sum_{t=1}^{T}\sum_{j=1}^{J}
    \bigl\lVert
      \tilde{\mathbf p}_{t,j}-\tilde{\hat{\mathbf p}}_{t,j}
    \bigr\rVert_2 .
  \label{eq:supp-metric-mpjpe}
\end{equation}

\paragraph{Root Position and Orientation Error.}
These metrics measure global anchor tracking. The orientation error uses the
quaternion geodesic angle
$\angle(q,\hat q)=2\arccos\!\bigl(\lvert\langle q,\hat q\rangle\rvert\bigr)$:
\begin{align}
  E^{\textup{pos}}_{\textup{root}}
    &=\frac{1}{T}\sum_{t=1}^{T}
      \bigl\lVert
        \mathbf p^{\textup{root}}_{t}
        -\hat{\mathbf p}^{\textup{root}}_{t}
      \bigr\rVert_2 ,
  \label{eq:supp-metric-rootpos}\\
  E^{\textup{rot}}_{\textup{root}}
    &=\frac{1}{T}\sum_{t=1}^{T}
      \angle\!\bigl(
        \mathbf q^{\textup{root}}_{t},
        \hat{\mathbf q}^{\textup{root}}_{t}
      \bigr).
  \label{eq:supp-metric-rootrot}
\end{align}

\paragraph{Root Velocity Error.}
This metric is the framewise body-frame L2 error of anchor linear velocity:
\begin{equation}
  E^{\textup{vel}}_{\textup{root}}
    =\frac{1}{T}\sum_{t=1}^{T}
      \bigl\lVert
        \mathbf v^{\textup{root}}_{t}
        -\hat{\mathbf v}^{\textup{root}}_{t}
      \bigr\rVert_2 .
  \label{eq:supp-metric-rootvel}
\end{equation}

\subsubsection{Comparison with Recent Motion Trackers}
\label{sec:exp-tracker-eval}

\begin{table*}[t]
  \centering
  \small
  \setlength{\tabcolsep}{2.5pt}
  \caption{Motion-tracking comparison on the validation set, reported
    separately for clean retargeted mocap references and the noisier PICO
    teleoperation references. Best per column in \textbf{bold}.}
  \label{tab:tracker-results}
  \begin{tabular}{@{}lcccccccccc@{}}
    \toprule
    & \multicolumn{5}{c}{\textbf{Mocap validation}}
    & \multicolumn{5}{c}{\textbf{PICO validation}} \\
    \cmidrule(lr){2-6}\cmidrule(lr){7-11}
    \textbf{Method}
      & SR\,$\uparrow$ & MPJPE\,$\downarrow$ & R.\ pos.\,$\downarrow$
      & R.\ ori.\,$\downarrow$ & R.\ vel.\,$\downarrow$
      & SR\,$\uparrow$ & MPJPE\,$\downarrow$ & R.\ pos.\,$\downarrow$
      & R.\ ori.\,$\downarrow$ & R.\ vel.\,$\downarrow$ \\
      & (\%) & (m) & (m) & (rad) & (m/s)
      & (\%) & (m) & (m) & (rad) & (m/s) \\
    \midrule
    TWIST2~\citep{ze2025twist2scalableportableholistic}
      & 43.1 & 0.248 & 0.372 & 0.235 & 0.245
      & 64.2 & 0.209 & 1.003 & 0.292 & 0.417 \\
    SONIC~\citep{luo2026sonicsupersizingmotiontracking}
      & 75.7 & 0.037 & 0.286 & \textbf{0.108} & 0.224
      & 82.1 & 0.038 & 0.575 & 0.104 & 0.272 \\
    HoloMotion~\citep{chen2026holomotion1technicalreport}
      & 64.6 & \textbf{0.034} & \textbf{0.103} & 0.227 & \textbf{0.205}
      & 97.0 & 0.035 & \textbf{0.127} & 0.244 & 0.273 \\
    \method{}
      & \textbf{91.7} & 0.040 & 0.260 & 0.109 & 0.228
      & \textbf{100.0} & \textbf{0.029} & 0.523 & \textbf{0.089} & \textbf{0.262} \\
    \bottomrule
  \end{tabular}
\end{table*}

\paragraph{Results.}
\method{} attains the highest SR on both subsets, with $91.7\%$ on mocap and
$100.0\%$ on PICO. On the PICO subset, it also obtains the lowest MPJPE, root
orientation error, and root velocity error. HoloMotion has the lowest root
position error on both subsets, while SONIC has a slightly lower MPJPE on the
mocap subset. These results indicate that \method{} provides a favorable
balance between tracking fidelity and stability on PICO references.

\subsubsection{Ablation Studies}

We ablate the two components that distinguish the tracker from a standard PPO
controller: (i)~\emph{w/o history encoder}, which removes the temporal Conv1d
encoder and feeds only the current-frame observation; and (ii)~\emph{w/o
rewind sampling}, which replaces failure-aware rewind with uniform clip
resampling ($\text{rewind\_prob}=0$).

All three variants use a shared reduced training setting: four GPUs,
$4096$ environments per GPU ($16\,384$ total), and $20$k iterations. The data,
PPO hyperparameters, and evaluation protocol are held fixed across the
variants, which are evaluated on the mocap validation subset.
Table~\ref{tab:tracker-ablation} reports their within-setting differences;
Table~\ref{tab:tracker-results} separately reports the final
$8$-GPU/$40$k checkpoint.

\begin{table}[t]
  \centering
  \small
  \setlength{\tabcolsep}{3.5pt}
  \caption{Tracker ablations on the mocap validation subset. All rows use a
    shared reduced setting of four GPUs, $4096$ environments/GPU, and $20$k
    iterations. Table~\ref{tab:tracker-results} reports the final checkpoint.
    Best per column in \textbf{bold}.}
  \label{tab:tracker-ablation}
  \begin{tabular}{@{}lccccc@{}}
    \toprule
    \textbf{Variant}
      & SR\,$\uparrow$ & MPJPE\,$\downarrow$ & R.\ pos.\,$\downarrow$
      & R.\ ori.\,$\downarrow$ & R.\ vel.\,$\downarrow$ \\
      & (\%) & (m) & (m) & (rad) & (m/s) \\
    \midrule
    Full (reduced)
      & \textbf{74.0} & 0.045 & 0.356 & \textbf{0.118} & 0.266 \\
    w/o rewind
      & 72.9 & \textbf{0.042} & \textbf{0.305} & 0.124 & \textbf{0.235} \\
    w/o history
      & 73.5 & 0.046 & 0.356 & 0.124 & 0.263 \\
    \bottomrule
  \end{tabular}
\end{table}

\paragraph{Results.}
The full configuration attains the highest success rate ($74.0\%$). Removing
the history encoder reduces both tracking fidelity and success. Removing rewind
sampling reduces success and root-orientation accuracy, although some position
metrics improve slightly. Rewind repeatedly revisits failure-adjacent segments,
which also helps the policy learn difficult motions earlier in training.

\subsubsection{Real-Robot Tracking}
\paragraph{Whole-body tracking on hardware.}
Figure~\ref{fig:tracker-real} pairs the operator and robot across static poses,
running, sidestepping, turning, and large changes in base height. The same
policy tracks all motions without motion-specific switching. The kneel--stand
and sit--stand sequences further demonstrate continuous transitions through
intermediate configurations on the humanoid.

\teleopitTrackerHardwareFigure

\subsection{Dexterous Hand Retargeting}
\label{sec:exp-hand}
The hand evaluation tests two complementary claims: the shared objective
transfers human poses across different robot-hand morphologies, and its
distance and frame objectives target complementary fingertip-distance and
thumb-frame alignment errors.

\subsubsection{Quantitative Protocol and Metrics}
\paragraph{Evaluation protocol.}
All six robot hands are evaluated on the same right-hand PICO recording. The
recording lasts $25.8$ seconds and was captured at $80$~Hz. In the metrics
below, $T$ denotes the number of time samples. All hands share the objective
weights, activation thresholds, preprocessing parameters, filter, and SLSQP
settings. The distance target uses raw human distances with scale $1.0$.

\paragraph{Ablation protocol.}
Objective ablations use the LinkerHand L20 right hand and hold the input,
initialization, joint limits, remaining objectives, filter, and solver settings
fixed.

\paragraph{Timing protocol.}
We run an untimed warm-up pass before measurement. Solve time covers the SLSQP
call and excludes sensing, visualization, communication, and robot control.

\paragraph{Continuous pose metrics.}
Direction error averages the angular error over the five semantic
finger-base-to-tip rays shared by every hand:
\begin{equation}
  E_{\mathrm{dir}}
  =
  \frac{1}{5T}
  \sum_{t=1}^{T}\sum_{i=1}^{5}
  \arccos\!\left(
    \operatorname{clip}
    \left(\hat{\mathbf v}^{R}_{t,i}\mathbin{\cdot}
          \hat{\mathbf v}^{H}_{t,i},-1,1\right)
  \right).
  \label{eq:supp-hand-direction}
\end{equation}
Here, $\hat{\mathbf v}^{H}_{t,i}$ and $\hat{\mathbf v}^{R}_{t,i}$ are unit
human and robot rays. Using a common set of semantic rays prevents hands with
different internal objective topologies from changing the evaluation
dimension. We report $E_{\mathrm{dir}}$ in degrees.

Normalized semantic keypoint error compares the five fingertips after
translation by the wrist or palm base and normalization by the corresponding
wrist-to-middle-base or palm-base-to-middle-base scale:
\begin{equation}
  E_{\mathrm{key}}
  =
  \frac{1}{5T}
  \sum_{t=1}^{T}\sum_{k=1}^{5}
  \left\|
    \frac{\mathbf p^{R}_{t,k}-\mathbf p^{R}_{t,w}}{s_R}
    -
    \frac{\mathbf p^{H}_{t,k}-\mathbf p^{H}_{t,w}}{s_H}
  \right\|_2 .
  \label{eq:supp-hand-keypoint}
\end{equation}

Site-distance tracking error averages the absolute error between human and
robot thumb-to-fingertip distances for the other four fingertips
$\mathcal P=\{\mathrm{index},\mathrm{middle},\mathrm{ring},\mathrm{little}\}$:
\begin{align}
  d^{X}_{t,p}
  &=
  \left\|\mathbf p^{X}_{t,\mathrm{thumb}}-\mathbf p^{X}_{t,p}\right\|_2,
  \quad X\in\{H,R\},
  \label{eq:supp-hand-site-distance}\\
  E_{\mathrm{dist}}
  &=
  \frac{1}{4T}
  \sum_{t=1}^{T}\sum_{p\in\mathcal P}
  \left|d^{R}_{t,p}-d^{H}_{t,p}\right|.
  \label{eq:supp-hand-distance}
\end{align}

Thumb-frame error is the geodesic distance between human and robot rotation
matrices constructed from the thumb-base primary and secondary axes:
\begin{equation}
  \begin{aligned}
    c_t
    &=
    \frac{
      \operatorname{tr}
      \left((\mathbf R^{R}_{t})^\top\mathbf R^{H}_{t}\right)-1
    }{2}, \\
    e_{\mathrm{frame},t}
    &=
    \arccos\!\left(\operatorname{clip}\!\left(c_t,-1,1\right)\right), \\
    E_{\mathrm{frame}}
    &=
    \frac{1}{T}\sum_{t=1}^{T}e_{\mathrm{frame},t}.
  \end{aligned}
  \label{eq:supp-hand-frame}
\end{equation}
We report $E_{\mathrm{frame}}$ in degrees. Mean solve time is the average
duration of the SLSQP call over the recording.

\subsubsection{Pose Fidelity and Cross-Embodiment Transfer}
\paragraph{Pose fidelity.}
Figure~\ref{fig:dex-multipose} retargets a range of operator hand poses onto a
single dexterous hand.
Across free-hand gestures, pinches, and opposed-thumb grasps, the solved
configuration preserves the finger directions and follows fingertip closure
without per-pose tuning.

\teleopitHandPoseFigure

\paragraph{Cross-embodiment transfer.}
Figure~\ref{fig:dex-multihand} applies the same objective form to a library of
commercial dexterous hands. Their proportions and thumb layouts differ
substantially. The normalized-direction formulation transfers across these
hands without per-vector human-to-robot scale tuning.

\teleopitHandTransferFigure

\begin{table}[t]
  \centering
  \footnotesize
  \setlength{\tabcolsep}{2pt}
  \caption{Cross-hand evaluation with shared objective and solver parameters.
    Active joints are independently optimized; passive joints follow active
    joints through fixed mimic couplings. All error and time metrics are
    lower-is-better.}
  \label{tab:supp-hand-cross}
  \begin{tabular}{@{}lrrrrrr@{}}
    \toprule
    \textbf{Hand}
      & \textbf{Active} & \textbf{Passive}
      & $\mathbf{E_{\mathrm{dir}}}$ & $\mathbf{E_{\mathrm{key}}}$
      & $\mathbf{E_{\mathrm{dist}}}$ & \textbf{Mean solve} \\
      & & & ($^\circ$) & & (mm) & (ms) \\
    \midrule
    Dex5           & 20 &  0 &  9.08 & 1.42 &  9.46 & 3.13 \\
    Inspire DFQ    &  6 &  6 & 29.53 & 1.98 & 14.45 & 1.56 \\
    LinkerHand L20 & 16 &  5 & 14.67 & 1.83 & 19.82 & 4.27 \\
    LinkerHand L6  &  6 &  5 & 19.43 & 1.74 & 21.15 & 1.93 \\
    Rohand         &  6 & 19 & 43.04 & 1.63 & 14.57 & 8.53 \\
    Sharpa Wave    & 22 &  0 &  9.02 & 1.40 & 18.05 & 4.04 \\
    \bottomrule
  \end{tabular}
\end{table}

\paragraph{Quantitative cross-hand results.}
Table~\ref{tab:supp-hand-cross} evaluates the shared numerical settings on the
full recording for each hand. The same objective and solver parameters run
across all six morphologies despite their different active and passive joint
counts. Mean solve time ranges from $1.56$ to $8.53$~ms.

Cross-hand accuracy is not uniform. Direction error ranges from
$9.02^\circ$ to $43.04^\circ$, while mean site-distance tracking error ranges
from $9.46$ to $21.15$~mm. The experiment therefore supports shared
parameterization across the six tested morphologies, while showing that
retargeting accuracy remains morphology dependent.

\subsubsection{Objective Ablations}
Figure~\ref{fig:dex-ablation} isolates the two secondary objectives on
representative poses.
Removing the distance objective weakens fingertip closure, while removing the
thumb-base frame objective weakens opposition roll. In a shared-parameter evaluation, the two objectives reduce their respective
distance and frame errors by $59.1\%$ and $29.2\%$.

\teleopitHandAblationFigure

\begin{table}[t]
  \centering
  \small
  \setlength{\tabcolsep}{3.5pt}
  \caption{Distance-objective ablation on LinkerHand L20.
    All metrics are lower-is-better; best per column in \textbf{bold}.}
  \label{tab:supp-hand-distance}
  \begin{tabular}{@{}lrrrr@{}}
    \toprule
    \textbf{Variant}
      & $\mathbf{E_{\mathrm{dist}}}$ & $\mathbf{E_{\mathrm{dir}}}$
      & $\mathbf{E_{\mathrm{key}}}$ & \textbf{Mean solve} \\
      & (mm) & ($^\circ$) & & (ms) \\
    \midrule
    w/o distance
      & 48.46 & \textbf{11.89} & \textbf{1.80}
      & \textbf{2.93} \\
    Full
      & \textbf{19.82} & 14.67 & 1.83
      & 4.27 \\
    \bottomrule
  \end{tabular}
\end{table}

\begin{table}[t]
  \centering
  \small
  \setlength{\tabcolsep}{3.5pt}
  \caption{Thumb-frame-objective ablation on LinkerHand L20.
    All metrics are lower-is-better; best per column in \textbf{bold}.}
  \label{tab:supp-hand-frame}
  \begin{tabular}{@{}lrrrr@{}}
    \toprule
    \textbf{Variant}
      & $\mathbf{E_{\mathrm{frame}}}$ & $\mathbf{E_{\mathrm{dir}}}$
      & $\mathbf{E_{\mathrm{key}}}$ & \textbf{Mean solve} \\
      & ($^\circ$) & ($^\circ$) & & (ms) \\
    \midrule
    w/o frame
      & 31.53 & 15.05 & 1.86 & \textbf{3.45} \\
    Full
      & \textbf{22.33} & \textbf{14.67} & \textbf{1.83} & 4.27 \\
    \bottomrule
  \end{tabular}
\end{table}

\paragraph{Distance objective.}
Table~\ref{tab:supp-hand-distance} shows that the distance objective reduces
mean site-distance tracking error from $48.46$ to $19.82$~mm, a $59.1\%$
reduction. This improvement trades off against a $2.78^\circ$ increase in
direction error, a $0.03$ increase in normalized keypoint error, and a
$1.34$~ms increase in mean solve time. The result shows that the distance
objective improves continuous fingertip-distance tracking under the fixed
ablation protocol.

\paragraph{Thumb-frame objective.}
Table~\ref{tab:supp-hand-frame} shows that the frame objective reduces mean
thumb-frame error from $31.53^\circ$ to $22.33^\circ$, a $9.20^\circ$ or
$29.2\%$ reduction. Direction and normalized keypoint errors also decrease by
$0.38^\circ$ and $0.03$, respectively, while mean solve time increases by
$0.82$~ms.

\subsection{Integrated Real-Robot Teleoperation}
\label{sec:exp-system}

We finally verify that the evaluated components operate together on the
humanoid during coordinated loco-manipulation.

\subsubsection{End-to-End Latency}

\begin{figure*}[t]
  \centering
  \includegraphics[width=\textwidth]{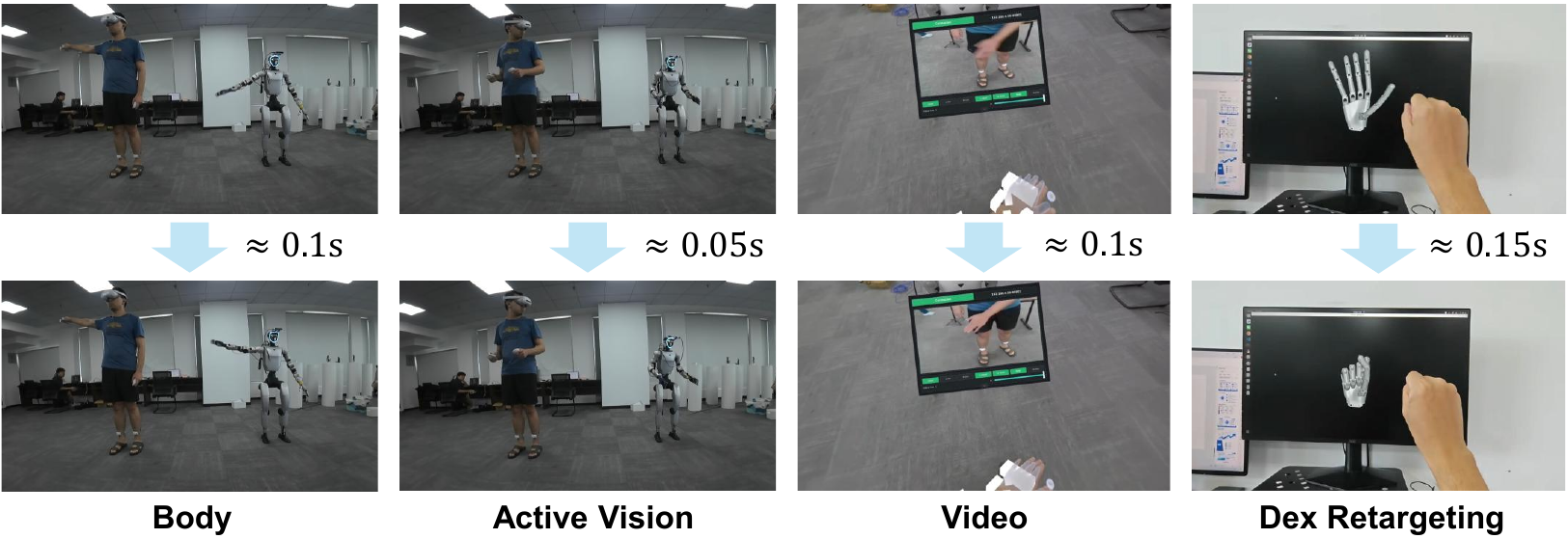}
  \caption{\textbf{Video-based estimates of end-to-end latency.} Paired images
    show the initiating event (top) and the downstream arrival (bottom).
    The estimated delays are approximately $0.10$~s for whole-body response,
    $0.05$~s for viewpoint-control response, $0.10$~s for the video stream
    arriving at the PICO headset, and $0.15$~s from human-hand motion to the
    displayed retargeted dexterous-hand configuration.}
  \label{fig:supp-delay}
\end{figure*}

We estimate latency from paired arrival events visible in the same video
recording. All network communication during the four measurements uses Wi-Fi.
For each path, latency is the elapsed time from the initiating motion to the
first visible downstream response. This procedure measures the complete path
to the output shown in Figure~\ref{fig:supp-delay}, including the sensing,
communication, control, and rendering contributions along that path.

The four measured paths respond within approximately $0.05$--$0.15$~s under
the recorded conditions. Timing uncertainty is bounded by the video sampling
interval and manual labeling of the first visible response. For dexterous
retargeting, latency ends when the retargeted hand configuration first appears
on the display.

\subsubsection{Closed-Loop Viewpoint Control}

The head-driven module provides live yaw--pitch viewpoint adjustment while the
operator continues to control the body and hands.

\begin{figure}[t]
  \centering
  \includegraphics[width=\linewidth]{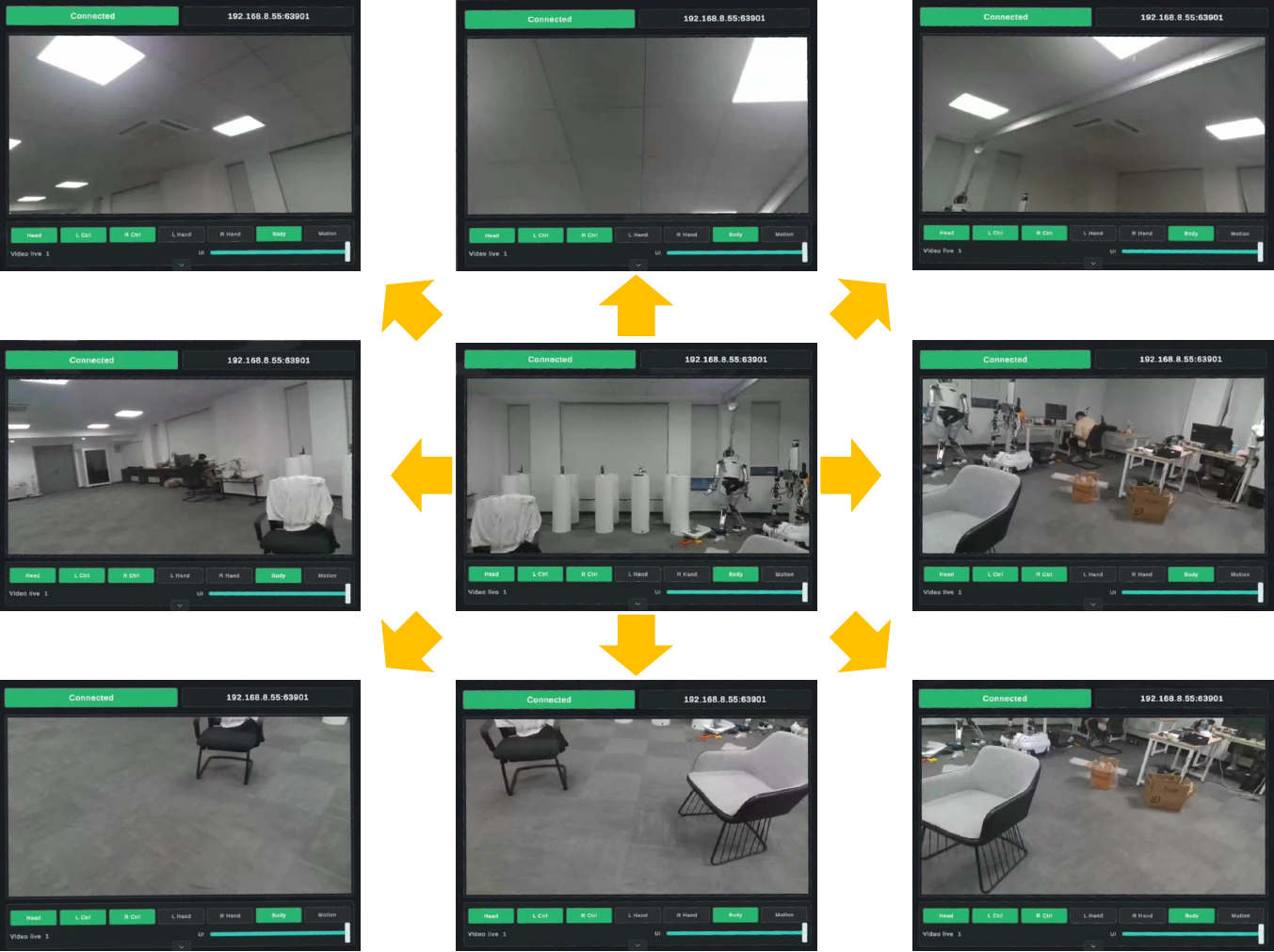}
  \caption{\textbf{Head-driven viewpoint control.} Views returned to the
    headset as the operator looks around from the neutral view (center). The
    eight surrounding frames show the camera coverage obtained by commanding
    yaw and pitch through head motion.}
  \label{fig:supp-active-vision-result}
\end{figure}

Figure~\ref{fig:active-vision-hardware} shows the yaw--pitch hardware and
summarizes its closed-loop function. Figure~\ref{fig:supp-active-vision-result} provides the corresponding
camera observations across the tested workspace. The neutral view appears at
the center, while the surrounding images cover vertical, horizontal, and
diagonal head commands. The diagonal views show simultaneous yaw and pitch
motion.

\subsubsection{Integrated Loco-Manipulation}

Figure~\ref{fig:integrated-teleoperation} exercises the complete sensing and
control path rather than any component in isolation. The demonstrations span
tabletop organization, transferring objects with a bag, operating lights and
doors, and sorting shelf items. They require the operator to coordinate base
placement and posture with arm reach, hand closure, and viewpoint selection,
showing that the asynchronous multi-rate runtime preserves the coupling needed
for extended whole-body tasks.

\teleopitIntegratedFigure

\subsection{Policy Learning from Teleopit Demonstrations}
\label{sec:supp-il}

We examine whether demonstrations recorded with Teleopit can directly support
autonomous whole body manipulation. We train ACT~\citep{zhao2023learningfinegrained}
and GR00T N1.7~\citep{nvidia2025gr00tn1openfoundation} on the same recordings.
During autonomous execution, the learned policy replaces the human operator at
the reference motion interface, while the motion tracker and robot control
stack remain unchanged. This design tests both the recorded data and the
interface through which task policies control the complete robot.

\subsubsection{Action Space and Control Interface}
\label{sec:supp-il-interface}

\paragraph{Design motivation.}
A task policy must coordinate the floating base, body joints, dexterous hands,
and active viewpoint. Predicting low level motor targets would additionally
require the task policy to learn balance and dynamic tracking from only a small
task dataset. Teleopit already solves this problem by converting a kinematic
whole body reference into feasible robot motion. We therefore train the task
policy to predict the same reference quantities produced during
teleoperation. The motion tracker remains responsible for whole body tracking,
whereas the hand and neck joint references are handled by their respective
controllers. The action space consequently preserves the control hierarchy
used to record the demonstrations.

\paragraph{Observation and recorded reference.}
At step $t$, the policy receives an RGB image $\mathbf I_t$ and the measured
robot state
\begin{equation}
  \mathbf s_t =
  [\mathbf q^{\mathrm{body}}_t,
   \mathbf q^{\mathrm{hand}}_t,
   \mathbf q^{\mathrm{neck}}_t]
  \in \mathbb R^{43},
  \label{eq:supp-il-state}
\end{equation}
which contains 29 body joint positions, 12 dexterous hand joint positions, and
two neck joint positions. Each recorded action is a 50D reference
\begin{equation}
  \mathbf a^{\mathrm{ref}}_t =
  [\mathbf p^{\mathrm{root}}_t,
   \mathbf r^{\mathrm{root}}_t,
   \bar{\mathbf q}^{\mathrm{body}}_t,
   \bar{\mathbf q}^{\mathrm{hand}}_t,
   \bar{\mathbf q}^{\mathrm{neck}}_t]
  \in \mathbb R^{50}.
  \label{eq:supp-il-canonical-action}
\end{equation}
The first two terms specify the 3D root position and 4D root orientation. The
remaining terms specify 29 body, 12 hand, and two neck joint references. The
root pose and body joints define the reference trajectory for the motion
tracker; the hand and neck terms preserve the commands recorded for the
other two robot subsystems.

\paragraph{Policy action.}
Global planar position and heading depend on where the robot starts and do not
describe the task itself. Learning them as absolute values would introduce
irrelevant variation across demonstrations. We instead express root motion
relative to the robot pose at the beginning of each predicted chunk. For a
future step $t+k$,
\begin{align}
  \Delta\mathbf p^{xy}_{t,k}
  &=
  \mathbf R_z(-\psi_t)
  (\mathbf p^{xy}_{t+k}-\mathbf p^{xy}_t),
  \label{eq:supp-il-root-xy}\\
  \Delta\mathbf R_{t,k}
  &=
  \mathbf R_t^\top\mathbf R_{t+k}.
  \label{eq:supp-il-root-rotation}
\end{align}
The policy action contains the 2D planar displacement
$\Delta\mathbf p^{xy}_{t,k}$, absolute root height $p^z_{t+k}$, and the
continuous 6D representation of $\Delta\mathbf R_{t,k}$
from~\citet{zhou2019continuityrotationrepresentations}, followed by the 43 body,
hand, and neck joint references. This gives a 52D policy action. Root height
remains absolute because it is defined with respect to the ground plane.
Using one source pose for the complete chunk also preserves a consistent
relative trajectory across its predicted steps.

\begin{figure*}[t]
  \centering
  \includegraphics[width=\textwidth]{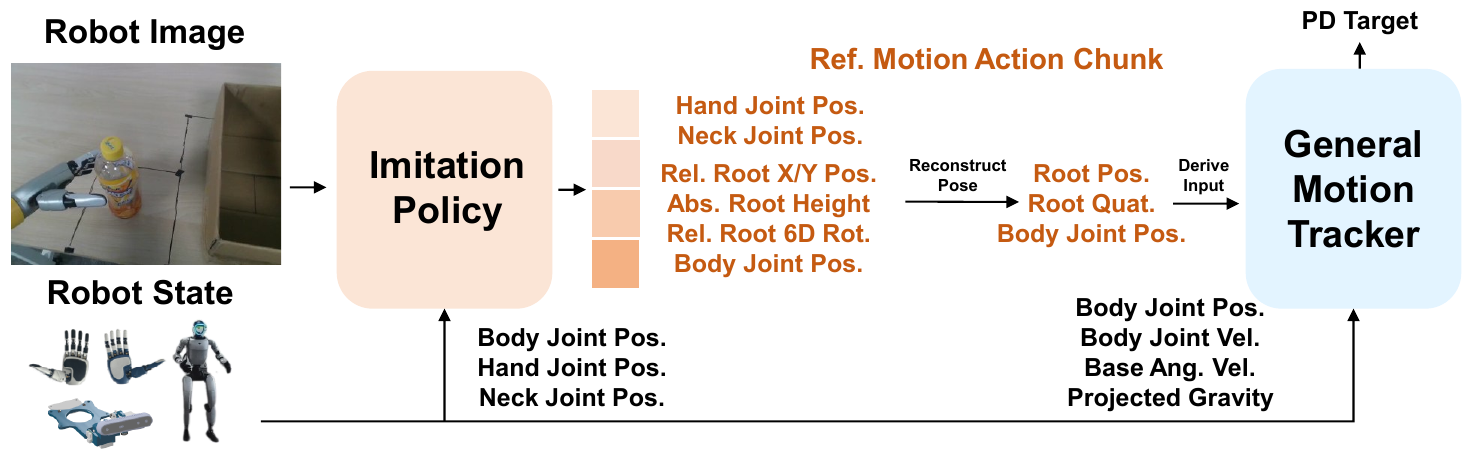}
  \caption{\textbf{Hierarchical interface for autonomous execution.}
    The task policy predicts a 52D reference motion chunk from the robot image
    and measured joint state. The initial root pose converts the relative root
    prediction back to a 50D robot reference. Root and body references enter
    the Teleopit motion tracker, which produces joint targets for the 200~Hz
    PD controller. Dexterous hand and neck joint references are sent to
    their corresponding controllers.}
  \label{fig:supp-il-pipeline}
\end{figure*}

Figure~\ref{fig:supp-il-pipeline} summarizes the deployment interface. The root
pose associated with the current observation converts the relative root action
back to the robot frame. The reconstructed body trajectory is supplied to the
50~Hz motion tracker, which produces joint targets tracked by the 200~Hz PD
controller. Hand and neck references bypass the motion tracker and are sent
directly to their corresponding controllers.

\begin{figure*}[t]
  \centering
  \includegraphics[width=\textwidth]{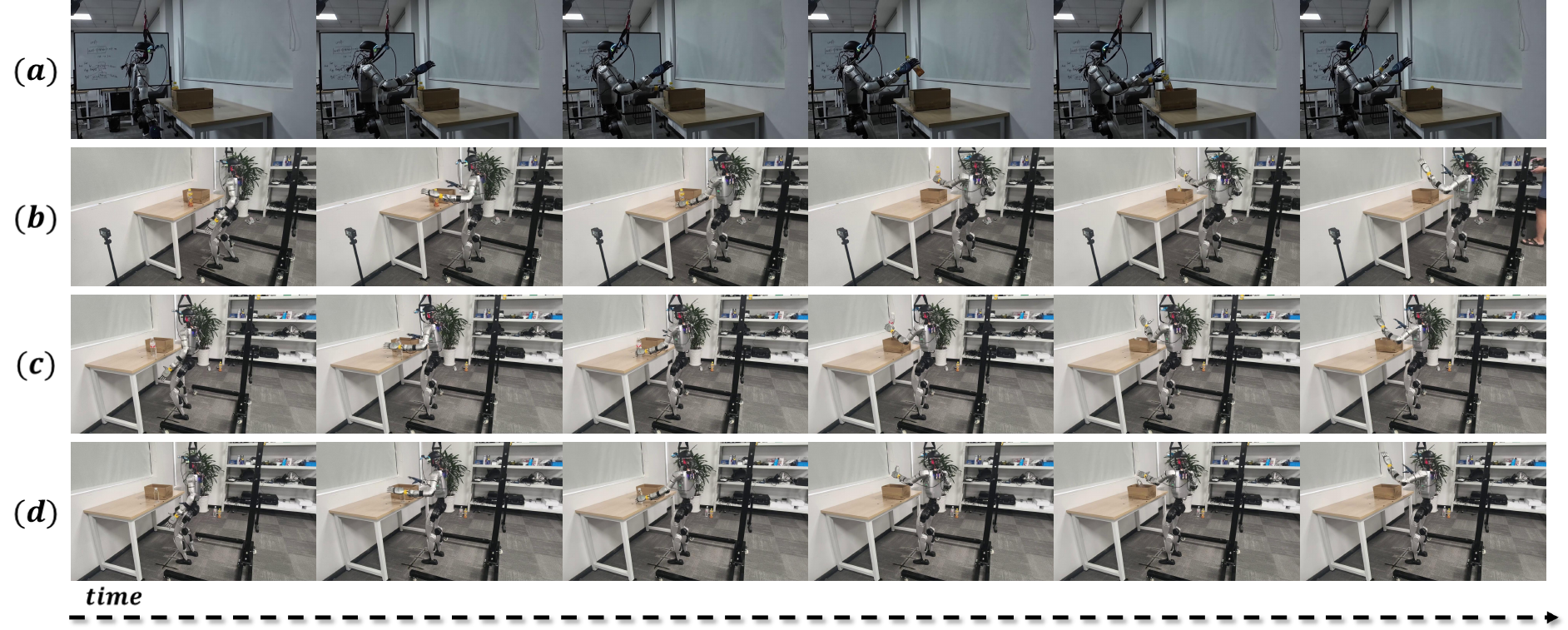}
  \caption{\textbf{Teleoperation collection and autonomous deployment.}
    Each row proceeds from left to right. (a)~Teleopit records an operator
    demonstration of grasping a bottle and placing it inside the adjacent box.
    (b)~GR00T N1.7 autonomously executes the same task. (c,d)~Without further
    demonstrations or fine tuning, the policy completes the task with two
    bottle types absent from its training demonstrations.}
  \label{fig:supp-il-collect-deploy}
\end{figure*}

\subsubsection{Training and Inference Setup}
\label{sec:supp-il-setup}

We convert the 96 successful 30~Hz recordings into a LeRobot
dataset~\citep{cadene2026lerobot} and train the two policies independently.
Both policies use the observation and action spaces defined above.
Table~\ref{tab:supp-il-training} reports the settings that affect optimization
or model capacity; data loading and checkpointing parameters are omitted.

\begin{table}[t]
  \centering
  \footnotesize
  \setlength{\tabcolsep}{3pt}
  \caption{Training settings used for ACT and GR00T N1.7. Batch size is
    reported per device.}
  \label{tab:supp-il-training}
  \begin{tabular}{@{}lcc@{}}
    \toprule
    \textbf{Setting} & \textbf{ACT} & \textbf{GR00T N1.7} \\
    \midrule
    Initialization & ImageNet R18 & N1.7 3B \\
    Optimizer updates & 150,000 & 20,000 \\
    Micro batch & 32 & 1 \\
    Accumulation steps & 1 & 8 \\
    Action chunk length & 50 & 40 \\
    Learning rate & $1\times10^{-5}$ & $1\times10^{-4}$ \\
    Weight decay & $1\times10^{-4}$ & $1\times10^{-5}$ \\
    Numeric precision & FP32 & BF16 \\
    \bottomrule
  \end{tabular}
\end{table}

ACT uses its ImageNet initialized ResNet18 visual encoder and is optimized in
full precision. For GR00T N1.7, the language and visual backbones remain
frozen, while the projector and diffusion model are optimized with BF16.
These choices preserve the pretrained representations of the
foundation model while adapting its action generation to the Teleopit
embodiment and reference space.

\begin{table}[t]
  \centering
  \small
  \setlength{\tabcolsep}{5pt}
  \caption{Inference and control settings shared by both policies.}
  \label{tab:supp-il-inference}
  \begin{tabular}{@{}ll@{}}
    \toprule
    \textbf{Setting} & \textbf{Value} \\
    \midrule
    Visual observation & $640\times480$ RGB at 30~Hz \\
    Policy action dimension & 52 \\
    Executed steps before replanning & 20 \\
    Motion tracker rate & 50~Hz \\
    Low level PD rate & 200~Hz \\
    \bottomrule
  \end{tabular}
\end{table}

As summarized in Table~\ref{tab:supp-il-inference}, we use asynchronous chunk
inference native to both policies. After executing 20 predicted steps, the
system acquires a new observation and replans. Policy inference runs
concurrently with robot control, so the current reference continues to drive
the motion tracker until the next chunk is available. ACT and GR00T therefore
differ in model architecture and prediction length but share the same
execution schedule and low level control stack.

\subsubsection{Data Collection and Autonomous Deployment}
\label{sec:supp-il-experiment}

\paragraph{Task and protocol.}
The task requires the robot to grasp a bottle from a table and place it inside
an adjacent box. We record 100 teleoperated trials at 30~Hz. A collection trial
is successful when the operator completes the task, and only successful trials
are included in the training set. We then evaluate each trained policy in 20
trials on the humanoid under the same task setup. A trial is counted as
successful if the bottle remains inside the box at the end of the rollout
without operator intervention.

\begin{table}[t]
  \centering
  \footnotesize
  \setlength{\tabcolsep}{3pt}
  \caption{Teleoperation collection yield and task success rates on the
    bottle-placement task. Both learned policies use the same 96
    demonstrations.}
  \label{tab:supp-il-results}
  \begin{tabular}{@{}lrrr@{}}
    \toprule
    \textbf{Control} & \textbf{Training set size} &
      \textbf{Success / trials} & \textbf{SR (\%)} \\
    \midrule
    Teleoperation & -- & 96 / 100 & 96.0 \\
    ACT & 96 & 18 / 20 & 90.0 \\
    GR00T N1.7 & 96 & 19 / 20 & 95.0 \\
    \bottomrule
  \end{tabular}
\end{table}

\paragraph{Quantitative results.}
Table~\ref{tab:supp-il-results} reports that teleoperation succeeds in 96 of
100 collection trials, yielding 96 demonstrations for policy learning. Trained
on this shared dataset, ACT succeeds in 18 of 20 autonomous trials and GR00T
N1.7 succeeds in 19 of 20. Thus, the same recordings and reference interface
support both a task specific action chunking policy and a vision language
action model at success rates of at least 90\%.

\paragraph{Qualitative results.}
Figure~\ref{fig:supp-il-collect-deploy}(a) presents the teleoperated
demonstration, while Figure~\ref{fig:supp-il-collect-deploy}(b) shows that the
learned policy reproduces its coordinated whole body reach, grasp, transport,
and placement. Figure~\ref{fig:supp-il-collect-deploy}(c,d) further shows
successful execution with bottles of different appearance and geometry,
without additional training. These examples qualitatively demonstrate
zero-shot transfer across bottle instances for the same task.

\section{Conclusion}

We presented \method{}, a full-embodiment humanoid teleoperation system that
uses VR to command whole-body motion, continuous dexterous hand motion, and
viewpoint control. The system integrates a history-aware motion tracker, an
optimization-based hand retargeter, an active vision module, and an
asynchronous runtime. Normalized finger directions allow the hand objective to
transfer across morphologies without hand-specific hyperparameter tuning, while
distance and thumb-frame objectives reduce complementary closure and alignment
errors.

The tracker achieves success rates of 91.7\% and 100.0\% on the mocap and PICO
validation subsets, respectively. Retargeting experiments cover more than a
dozen dexterous hands, and real-robot teleoperation combines locomotion,
continuous hand control, and viewpoint control. Demonstrations recorded with
\method{} also support downstream policy learning. Trained on a shared set of
96 demonstrations, ACT and GR00T N1.7 achieve task success rates of 90.0\% and
95.0\%, respectively, when deployed on the humanoid.

\clearpage
\bibliographystyle{unsrtnat}
\bibliography{references}

\end{document}